\documentclass[sigconf]{acmart}
\acmSubmissionID{593}

\usepackage{booktabs}
\usepackage{caption}
\usepackage{subcaption}
\usepackage{multirow}
\usepackage{enumitem}
\usepackage{fmtcount}
\usepackage{microtype}

\usepackage[ruled]{algorithm2e} 

\SetAlFnt{\small}
\SetAlCapFnt{\small}
\SetAlCapNameFnt{\small}
\SetAlCapHSkip{0pt}

\newcommand{\image}{I}
\newcommand{\inputcontent}{\image_c}
\newcommand{\inputstyle}{\image_s}

\newcommand{\augmentedimage}{\image^{+}}
\newcommand{\negativeexample}{\image^{-}}

\newcommand{\loss}{\mathcal{L}}
\newcommand{\latentcode}{\mathbf{z}}
\newcommand{\stylelatent}{\hat{\latentcode}}
\newcommand{\outputlatent}{\tilde{\latentcode}}

\newcommand{\generator}{G}

\usepackage{ifthen}

\newcommand{\warning}[1]{{\it\color{red} #1}}
\newcommand{\toremove}[1]{{\it\color{red} (To remove) #1}}
\newcommand{\note}[1]{{\it\color{blue} #1}}
\newcommand{\nothing}[1]{}

\definecolor{FanColor}{rgb}{0.8,0,0.8}

{

\renewcommand{\warning}[1]{}
\renewcommand{\toremove}[1]{}
\renewcommand{\note}[1]{}
\renewcommand{\nothing}[1]{}
}{}

\copyrightyear{2022} 
\acmYear{2022} 
\setcopyright{rightsretained} 
\acmConference[SIGGRAPH '22 Conference Proceedings]{Special Interest Group on Computer Graphics and Interactive Techniques Conference Proceedings}{August 7--11, 2022}{Vancouver, BC, Canada}
\acmBooktitle{Special Interest Group on Computer Graphics and Interactive Techniques Conference Proceedings (SIGGRAPH '22 Conference Proceedings), August 7--11, 2022, Vancouver, BC, Canada}
\acmDOI{10.1145/3528233.3530736}
\acmISBN{978-1-4503-9337-9/22/08}

\begin{document}

\title{Domain Enhanced Arbitrary Image Style Transfer via Contrastive Learning}

\author{Yuxin Zhang$^{1,2}$\hspace{0.2in} Fan Tang$^{3*}$\hspace{0.2in} Weiming Dong$^{1,2*}$\hspace{0.2in} Haibin Huang$^4$\\ Chongyang Ma$^4$\hspace{0.2in} Tong-Yee Lee$^5$\hspace{0.2in} Changsheng Xu$^{1,2}$}
\affiliation{
$^1$\institution{NLPR, Institute of Automation, Chinese Academy of Sciences \hspace{0.2in} $^2$School of Artificial Intelligence, UCAS\\
$^3$School of AI, Jilin University \hspace{0.15in} $^4$Kuaishou Technology \hspace{0.15in} $^5$National Cheng-Kung University}
\country{}
}

\renewcommand{\shortauthors}{Y. Zhang, F. Tang, W. Dong, H. Huang, C. Ma, T.-Y. Lee, and C. Xu}
\renewcommand{\authors}{Yuxin Zhang, Fan Tang, Weiming Dong, Haibin Huang, Chongyang Ma, Tong-Yee Lee, and Changsheng Xu}


\begin{abstract}
\footnotetext{* Co-corresponding authors. Emails: tangfan@jlu.edu.cn, weiming.dong@ia.ac.cn}
In this work, we tackle the challenging problem of arbitrary image style transfer using a novel style feature representation learning method.
A suitable style representation, as a key component in image stylization tasks, is essential to achieve satisfactory results.
Existing deep neural network based approaches achieve reasonable results with the guidance from second-order statistics such as Gram matrix of content features.
However, they do not leverage sufficient style information, which results in artifacts such as local distortions and style inconsistency.
To address these issues, we propose to learn style representation directly from image features instead of their second-order statistics, by analyzing the similarities and differences between multiple styles and considering the style distribution.
Specifically, we present Contrastive Arbitrary Style Transfer (CAST), which is a new style representation learning and style transfer method via contrastive learning.
Our framework consists of three key components, i.e., a multi-layer style projector for style code encoding, a domain enhancement module for effective learning of style distribution, and a generative network for image style transfer.
We conduct qualitative and quantitative evaluations comprehensively to demonstrate that our approach achieves significantly better results compared to those obtained via state-of-the-art methods.
Code and models are available at \url{https://github.com/zyxElsa/CAST_pytorch}.
\end{abstract}





\begin{teaserfigure}
\centering
\includegraphics[width=\linewidth]{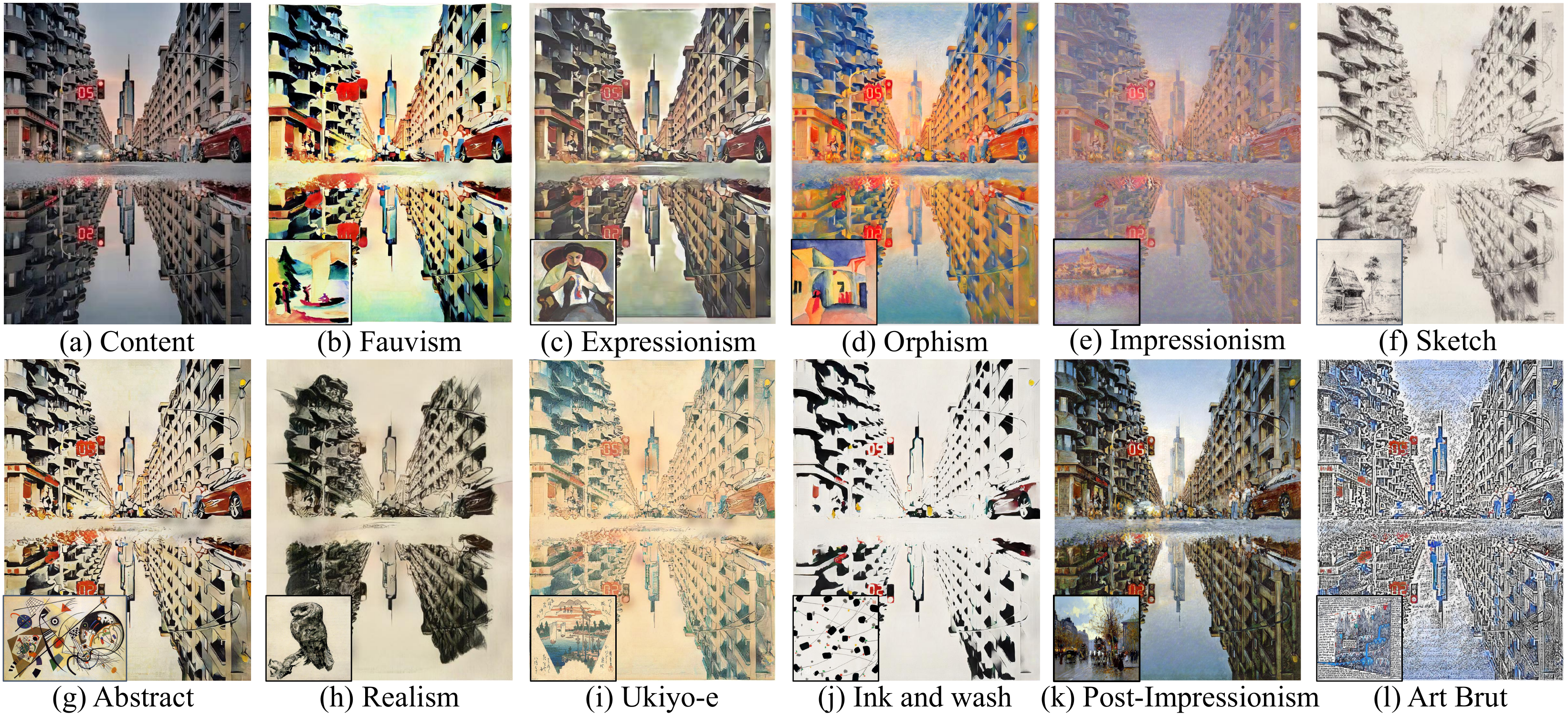}
 \vspace{-15pt}
\caption{Style transfer results using our method, which can robustly and effectively handle various painting styles.
The input content image is in the top-left corner and the style reference is shown as the inset for each result.
Our method can faithfully capture the style of each painting and generate a result with a unique artistic visual appearance.
Artists of style images: (b) August Macke, (c) August Macke, (d) August Macke, (e) Claude Monet, (f) Vincent van Gogh, (g) Wassily Kandinsky, (h) Vincent van Gogh, (i) Hiroshige, (j) Guanzhong Wu, (k) Edouard Cortes, and (l) Howard Finster.
}
\label{fig:teaser}
\end{teaserfigure}

\maketitle


\section{Introduction}
\label{sec:intro}


\newcommand{\castiestfigurewidth}{0.19}
\begin{figure}
\centering
\begin{subfigure}[t]{\castiestfigurewidth\linewidth}
\includegraphics[width=1\linewidth]{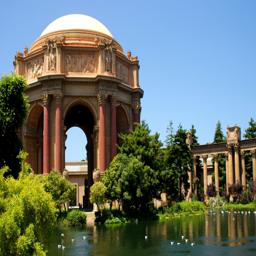}
\includegraphics[width=1\linewidth]{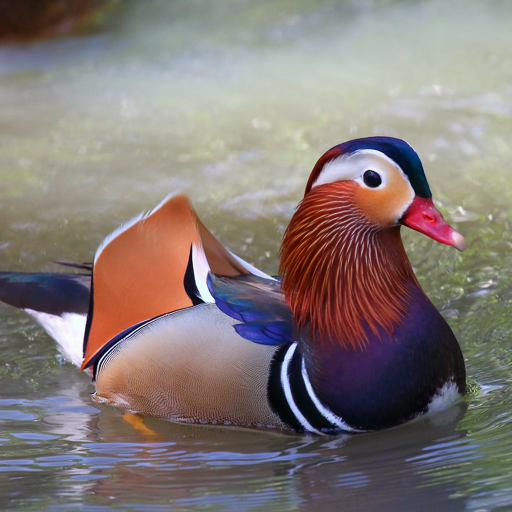}
\caption{Content}
\label{fig:IECAST_content}
\end{subfigure}
\begin{subfigure}[t]{\castiestfigurewidth\linewidth}
\includegraphics[width=1\linewidth]{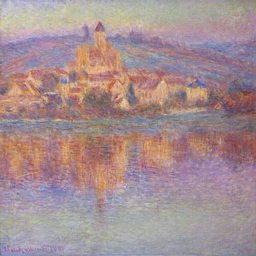}
\includegraphics[width=1\linewidth]{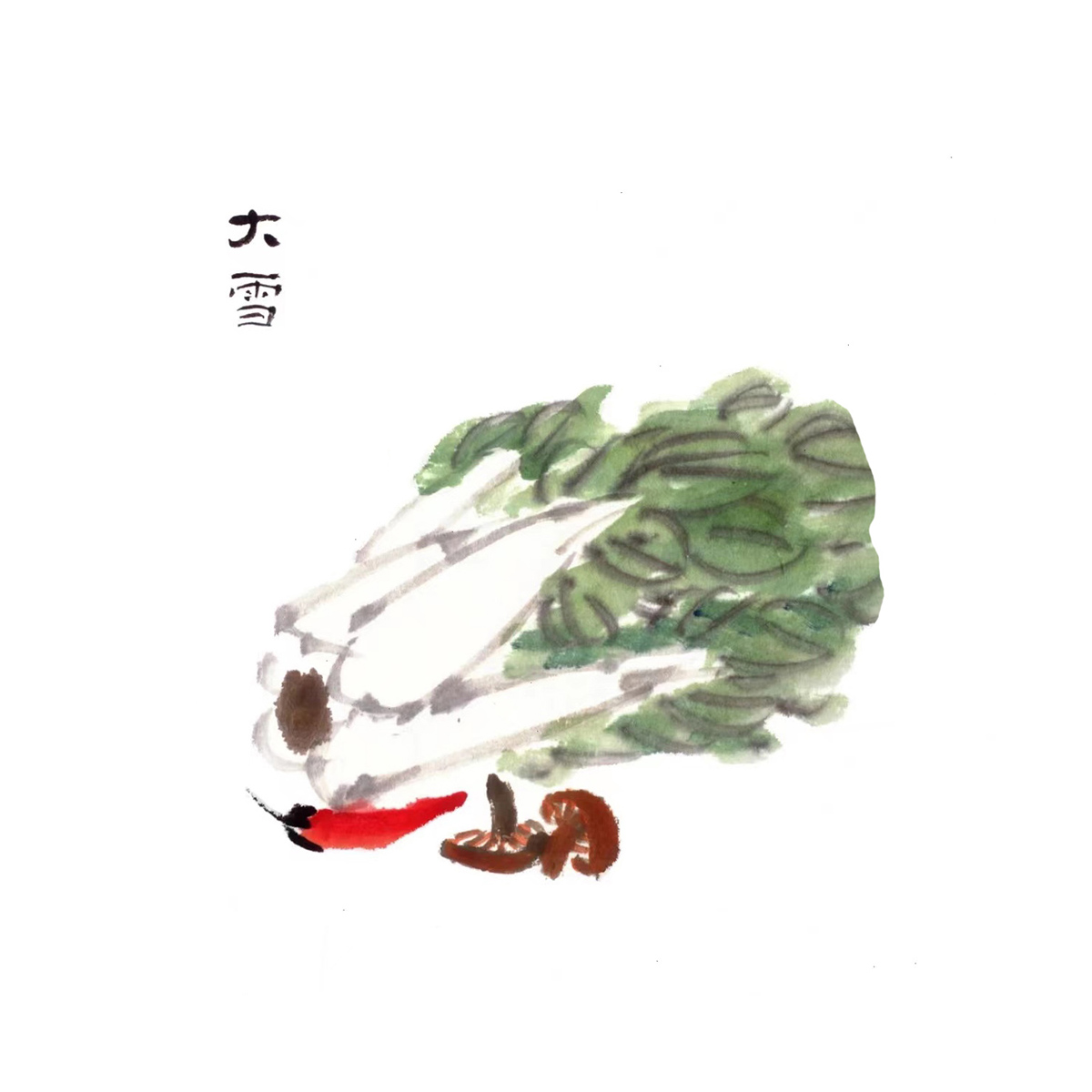}
\caption{Style}
\label{fig:IECAST_style}
\end{subfigure}
\begin{subfigure}[t]{\castiestfigurewidth\linewidth}
\includegraphics[width=1\linewidth]{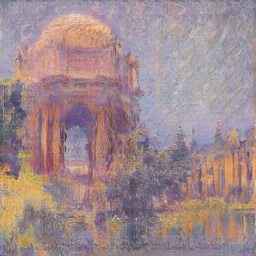}
\includegraphics[width=1\linewidth]{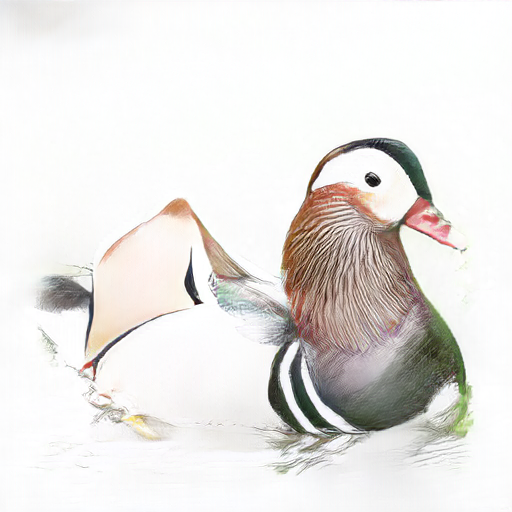}
\caption{Ours}
\label{fig:IECAST_ours}
\end{subfigure}
\begin{subfigure}[t]{\castiestfigurewidth\linewidth}
\includegraphics[width=1\linewidth]{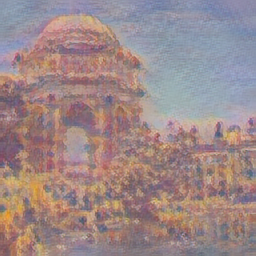}
\includegraphics[width=1\linewidth]{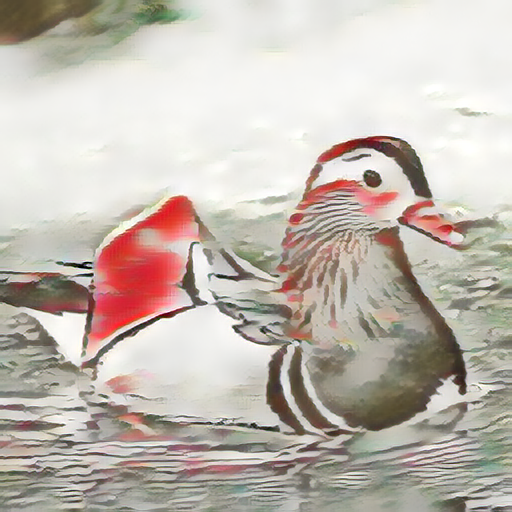}
\caption{AdaAttN}
\label{fig:IECAST_AdaAttN}
\end{subfigure}
\begin{subfigure}[t]{\castiestfigurewidth\linewidth}
\includegraphics[width=1\linewidth]{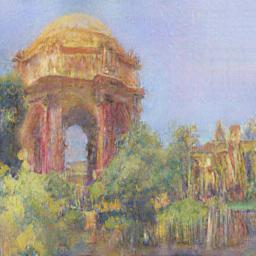}
\includegraphics[width=1\linewidth]{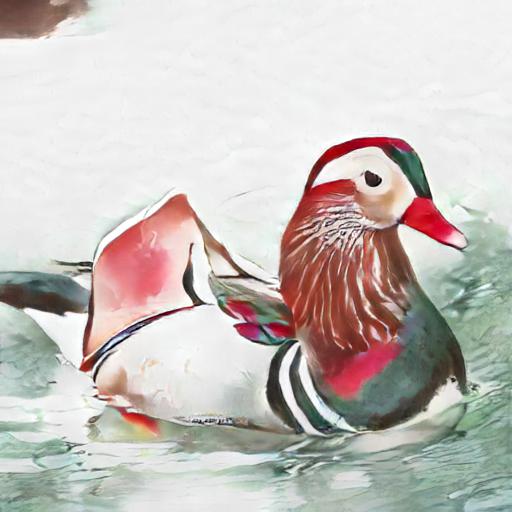}
\caption{IEST}
\label{fig:IECAST_theirs}
\end{subfigure}
\caption{Compared with AdaAttnN~\cite{liu2021adaattn} and IEST~\cite{chen2021artistic} which rely on second-order statistics, our method can faithfully transfer styles while ensuring structural consistency with the content images. Artists of style images: Claude Monet and Jueli Shi.
}
\label{fig:IECAST}
\end{figure}


If a picture is worth a thousand words, then an artwork tells the whole story.
Art styles, which describe the way the artwork looks, are the manner in which the artist portrays his or her subject matter and how the artist expresses his or her vision.
Style is determined by the characteristics that describe the artwork, such as the way the artist employs form, color, and composition.
Artistic style transfer, as an efficient way to create a new painting by combining the content of a natural images and the style of an existing painting image, is a major  research topic in computer graphics and computer vision~\cite{Liao:2017:VAT,Jing:2020:NSTReview}, with style representation as the most important issue.

Since Gatys et al.~\shortcite{Gatys:2016:IST} proposed to use Gram matrix as artistic style representation, high-quality visual results are generated by advanced neural style transfer networks.
Despite the remarkable progress made in the field of arbitrary image style transfer, the second-order feature statistics (Gram matrix or mean/variance) style representation has restricted the further development and application.
As shown in Figure~\ref{fig:teaser}, the appearances of different artwork styles vary considerably in terms of not only the colors and local textures but also the layouts and compositions.
Figures~\ref{fig:IECAST_AdaAttN} and~\ref{fig:IECAST_theirs} show the results of two recently proposed state-of-the-art style transfer approaches.
We obverse that aligning the distributions of neural activation between images using second-order statistics results in difficulty to capture the color distribution or the special layouts, or imitate specific detailed brush effects of different styles.

In this paper, we revisit the core problem for neural style transfer, that is, the proper artistic style representation.
The widely used second-order statistics as a global style descriptor can distinguish styles to some extent, but they are not the optimal way to represent styles.
By second-order statistics, arbitrary stylization formulates styles through artificially designed image features and loss functions in a heuristic manner.
In other words, the network learns to fit the second-order statistics of the style image and generated image, instead of the style itself.
Exploring the relationship and distribution of styles directly from artistic images instead of using pre-defined style representations is worthwhile.

Toward this end, we propose to improve arbitrary style transfer with a novel style representation by contrastive learning-based optimization.
Our key insight is that a person without artistic knowledge has difficulty defining the style if only one artistic image is given, but identifying the difference between different styles is relatively easy.
Specifically, we present a novel \textbf{C}ontrastive \textbf{A}rbitrary \textbf{S}tyle \textbf{T}ransfer (\textbf{CAST}) framework for image style representation and style transfer.
CAST consists of a backbone based on an encoder-transformation-decoder structure, a \textbf{m}ulti-layer \textbf{s}tyle \textbf{p}rojector (\textbf{MSP}) module, and a \textbf{d}omain \textbf{e}nhancement (\textbf{DE}) module.
We introduce contrastive learning to consider the positive and negative relationships between styles, and we use DE to learn the distribution of overall art image domains.
To capture the style features at various scales, our MSP module projects the features of each layer of the style image to the corresponding style encoding space.

Our contribution can be summarized as follows:
\begin{itemize}[leftmargin=*,topsep=1pt]
\setlength\itemsep{0pt}
\item We propose an MSP module for style encoding and a novel CAST model for encoder-transformation-decoder-based arbitrary style transfer without using the second-order statistics as style representations.

\item We introduce contrastive learning and domain enhancement by considering the relationships between positive and negative examples as well as the global distribution of styles, which solves the problem that existing style transfer models cannot fully utilize a large amount of style information. 

\item Experiments show that our method achieves state-of-the-art style transfer results in terms of visual quality.
A challenging subjective survey was conducted, as inspired by the Turing test, to show that output of CAST could mislead participants from telling the fake painting images from real ones.
\end{itemize}


\begin{figure*}
\centering
\includegraphics[width=\linewidth]{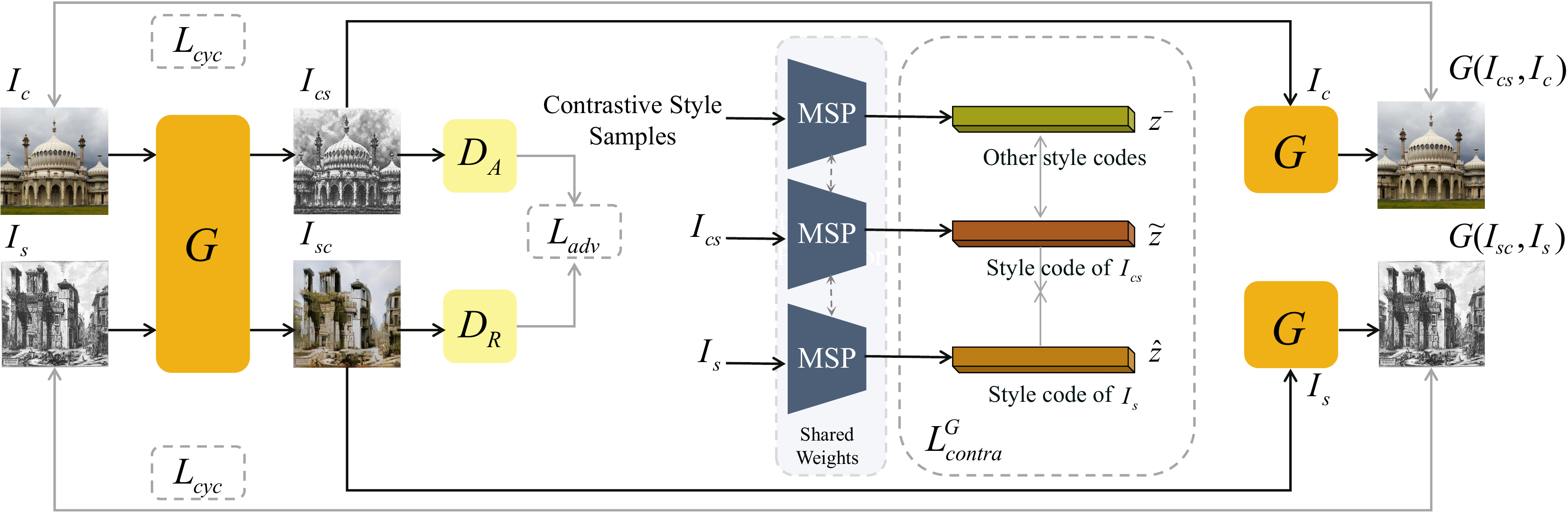}
\vspace{-15pt}
\caption{CAST consists of an encoder-transformation-decoder-based generator $\generator$, a multi-layer style projector (MSP) module, and a domain enhancement module.
We first generate images $\image_{cs}$ and $\image_{sc}$ from the content image $\inputcontent$ and the style image $\inputstyle$ using the  generator.
Then, $\image_{cs}$ and $\inputstyle$ are fed into the MSP module to generate the corresponding style code $\outputlatent$ and $\stylelatent$, which will be used as positive samples in the style contrastive learning process. The style codes $\latentcode^{-}$ of other artistic images in the style bank will be used as negative samples.
We compute a contrastive style loss $\loss_{contra}^{G}$ based on these style codes.
DE module is based on the adversarial loss $\loss_{adv}$ and the cycle consistency loss $\loss_{cyc}$.
Artist of style image: Giovanni Battista Piranesi.
}
\vspace{-10pt}
\label{fig:framework}
\end{figure*}




\section{Related Work}

\paragraph{Image style transfer.}
Traditional style transfer methods such as stroke-based rendering~\cite{Fivser:2016:Stylit} and image filtering~\cite{Wang:2004:EEP} typically use low-level hand-crafted features.
Gatys et al.~\shortcite{Gatys:2016:IST} and the follow-up variants \cite{Gatys:2017:CPF,Kolkin:2019:STR} demonstrate that the statistical distribution of features extracted from pre-trained deep convolutional neural networks can capture style patterns effectively.
Although the results are remarkable, these methods formulate the task as a complex optimization problem, which leads to high computational cost.
Some recent approaches rely on a learnable neural network to match the statistical information in feature space for efficiency.
Per-style-per-model methods~\cite{johnson2016perceptual,gao2020fast, puy2019flexible} train a specific network for each individual style.
Multiple-style-per-model methods~\cite{chen2017stylebank,zhang2018multi,dumoulin2016learned, ulyanov2016texture} represent multiple styles using one single model.

Arbitrary style transfer methods~\cite{Li:2017:UST,deng2020arbitrary,svoboda2020two,Wu:2021:Styleformer,deng2021stytr2} build more flexible feed-forward architectures to handle an arbitrary style using a unified model.
AdaIN~\cite{Huang:2017:AdaIn} and DIN~\cite{jing2020dynamic} directly align the overall statistics of content features with the statistics of style features and adopt conditional instance normalization.
However, dynamic generation of affine parameters in the instance normalization layer may cause distortion artifacts.
Instead, several methods follow the encoder-decoder manner, where feature transformation and/or fusion is introduced into an auto-encoder-based framework. For example, Li et al.~\shortcite{Li:2019:LLT} learn a cross-domain feature linear transformation matrix (LST) to enable universal style transfer and generate the desired stylization results by decoding from the transformed features.
Park et al.~\shortcite{Park:2019:AST} introduce SANet to flexibly match the semantically nearest style features onto the content features.
Deng et al.~\shortcite{Deng:2021:MCC} propose MCCNet to fuse exemplar style features and input content features by multi-channel correlation for efficient style transfer.
An et al.~\shortcite{An:2021:Artflow} propose reversible neural flows and an unbiased feature transfer module (ArtFlow) to prevent content leak during universal style transfer.
Liu et al.~\shortcite{liu2021adaattn} present an adaptive attention normalization module (AdaAttN) to consider both shallow and deep features for attention score calculation.
GAN-based methods~\cite{Zhu:2017:CycleGAN,svoboda2020two, kotovenko2019content_transformation, kotovenko2019content, sanakoyeu2018style} have been successfully used in collection style transfer, which considers style images in a collection as a domain~\cite{chen2021dualast, xu2021drb,Lin:2021:DAM}.

\paragraph{Contrastive learning.}
Contrastive learning has been used in many applications, such as image dehazing~\cite{Wu:2021:CLC}, context prediction~\cite{Cruz:2019:VPL}, geometric prediction~\cite{Liu:2019:EUD} and image translation.
Contrastive learning is introduced in image translation to preserve the content of the input~\cite{Han:2021:DCL} and reduce mode collapse~\cite{Liu:2021:DivCo,Jeong:2021:Contrad,Kang:2020:ContraGAN}.
CUT~\cite{park2020CUT} proposes patch-wise contrastive learning by cropping input and output images into patches and maximizing the mutual information between patches.
Following CUT, TUNIT~\cite{baek2021tunit} adopts contrastive learning on images with similar semantic structures.
However, the semantic similarity assumption does not hold for arbitrary style transfer tasks, which leads the learned style representations to a significant performance drop.
IEST~\cite{chen2021artistic} applies contrastive learning to image style transfer based on feature statistics (mean and standard deviation) as style priors.
The contrastive loss is calculated only within the generated results.
Contrastive learning in IEST is an auxiliary method to associate stylized images sharing the same style, and the ability comes from the feature statistics from pre-trained VGG.
Differently, we introduce contrastive learning for style representation by proposing a novel framework that uses visual features comprehensively to represent style for the task of arbitrary image style transfer.



\section{Method}
As shown in Figure~\ref{fig:framework}, our framework consists of three key components:
(1) a multi-layer style projector which is trained to project features of artistic image into style code;
(2) a contrastive style learning module which is applied to guide both the training of the multi-layer style projector and the style image generation;
and (3) a domain enhancement scheme to further help learn the distribution of artistic image domain.
All these components are used for learning style representations to measure the difference between the input artistic images and generated results and thus, they could be applied to different kinds of arbitrary style transfer networks.

\subsection{Multi-layer Style Projector}

Our goal is to develop an arbitrary style transfer framework that can capture and transfer the local stroke characteristics and overall appearance of an artistic image to a natural image.
A key component is to find a suitable style representation which can be used to distinguish different styles and further guide the generation of style images.
To this end, we design an MSP module, which includes a style feature extractor and a multi-layer projector.
Instead of using features from a specific layer or a fusion of multiple layers, our MSP projects features of different layers into separate latent style spaces to encode local and global style cues.

\begin{figure}
\centering
\includegraphics[width=\linewidth]{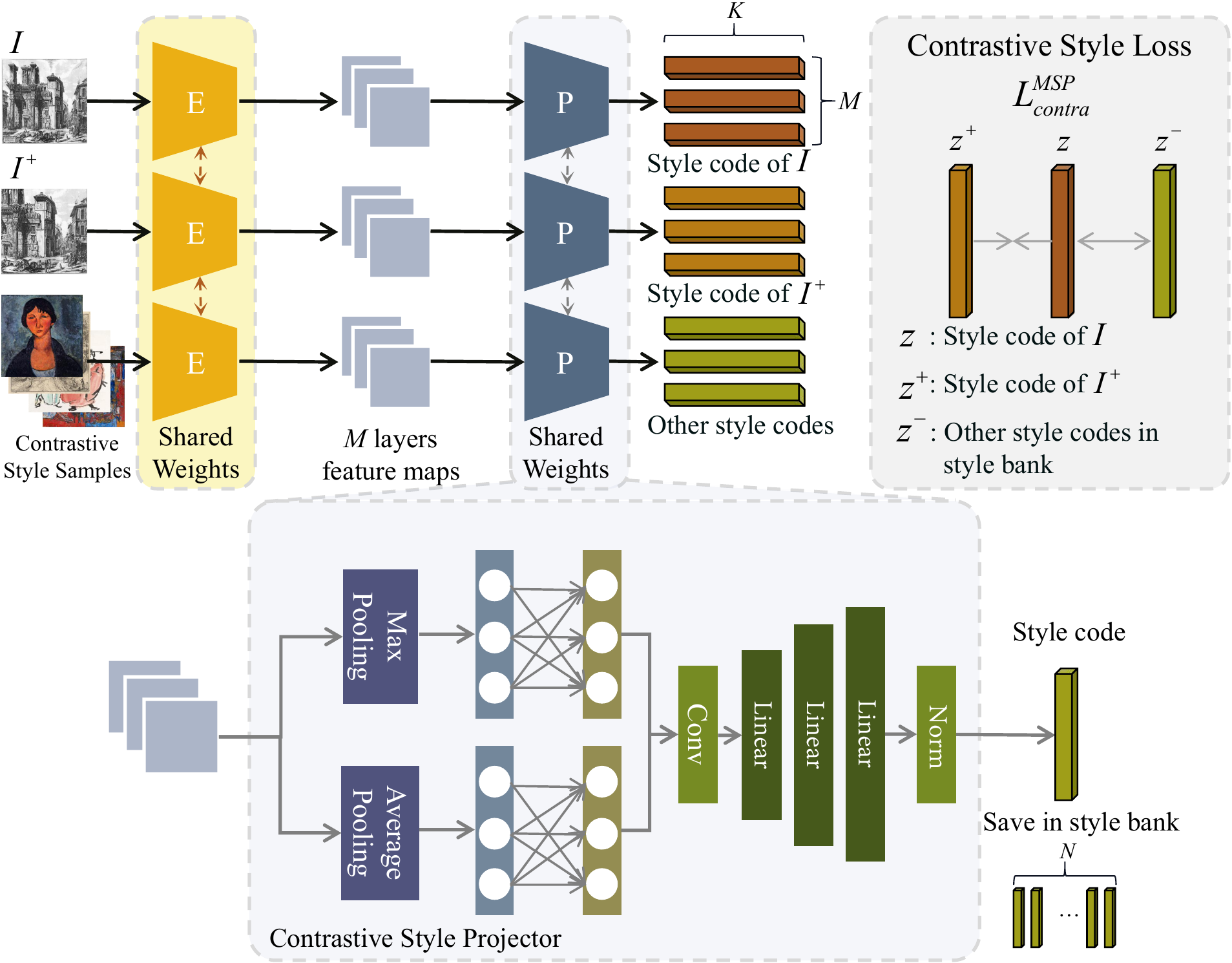}
\caption{Overview of our MSP module, which includes a VGG-19 based style feature extractor $E$ and a multi-layer projector $P$. 
$P$ maps the extracted features to style codes $\{ \latentcode \}$ which are saved in the style memory bank to compute the contrastive style loss $\loss_{contra}^{MSP}$.
Artists of style images: Giovanni Battista Piranesi and Amedeo Modigliani.
}
\label{fig:msp_overview}
\end{figure}

Specifically, we adapt VGG-19~\cite{simonyan2014very} and finetune the VGG-19 model pre-trained on ImageNet with a collection of 18,000 artistic images in 30 categories.
We then select $M$ layers of feature maps in VGG-19 as input to our multi-layer projector (we use layers of ReLU1\_2, ReLU2\_2, ReLU3\_3, and ReLU4\_3 in all experiments).
We use max pooling and average pooling to capture the mean and peak value of features.
The multi-layer projector consists of pooling, convolution, and several multilayer perceptron layers, and it projects the style features into a set of $K$-dimensional latent style code, as shown in Figure~\ref{fig:msp_overview}.

After training, MSP can encode an artistic image into a set of latent style code $\{ \latentcode_i | i \in [1, M], \latentcode_i \in \mathbb{R}^K\}$, which can be plugged into an existing style transfer network (i.e., replacing the mean and variance in AdaIN~\cite{Huang:2017:AdaIn}) as the guidance for stylization.
Next, we will describe how to jointly train MSP and style transfer networks with a contrastive learning strategy.

\subsection{Contrastive Style Learning}

As demonstrated above, the style code $\{ \latentcode_1, \latentcode_2,..., \latentcode_M \}$ of an image can be used as the target for MSP training and the guidance for the style transfer network.
However, we lack the ground-truth style code for supervised training.
Therefore, we adopt contrastive learning and design a new contrastive style loss as an implicit measurement for network training.

When train the MSP module, an image $\image$ and its augmented version $ \augmentedimage $ (random resizing, cropping, and rotations) are fed into a $M$-layer style feature extractor, which is the pre-trained VGG-19 network.
The extracted style features are then sent to the multi-layer projector, which is an $M$-layer neural network and maps the style features to a set of $K$-dimensional vectors $\{ \latentcode \}$.
The contrastive representation learns the visual styles of images by maximizing the mutual information between $\image$ and $\augmentedimage$ in contrast to other artistic images within the dataset considered as negative samples $\{ \negativeexample \}$.
Specifically, the images $\image$, $\augmentedimage$, and $N$ negative samples are respectively mapped into $M$ groups of $K$-dimensional vectors $\latentcode$, $\latentcode^{+} \in \mathbb{R}^K$ and $\{ \latentcode^{-} \in \mathbb{R}^K \}$. The vectors are normalized to prevent collapsing.
Following \cite{van2018representation}, we define the contrastive loss function to train our MSP module as:
\begin{equation}
\begin{aligned}
\loss_{contra}^{MSP}=-\sum_{i=1}^M{\log \frac{\exp(\latentcode_{i} \cdot {\latentcode_{i}^{+}}/ \tau)}{\exp (\latentcode_{i} \cdot {\latentcode_{i}^{+}} / \tau)+\sum_{j=1}^N{\exp(\latentcode_{i} \cdot {\latentcode_{i_{j}}^{-}} / \tau)}}},
\end{aligned}
\label{eqn:loss_msp}
\end{equation}
where $\cdot$ denotes the dot product of two vectors, and $\tau$ is a temperature scaling factor and is set to be $0.07$ in all of our experiments.
Meanwhile, we maintain a large dictionary of 4096 negative examples using a memory bank architecture following MOCO~\cite{He:2019:MOCO}.
It is worth noting that we calculate the contrastive loss between \emph{images}, as opposed to CUT~\cite{park2020CUT} which adopts contrastive learning by cropping images into patches and maximizing the mutual information between \emph{patches}.


The contrastive representation also provides proper guidance for the generator $\generator$ to transfer styles between images.
We adopt the same form of contrastive loss as used for learning MSP in Eq.~(\ref{eqn:loss_msp}), but compute the loss using the contrastive representations of the output image $\image_{cs}$ and the reference style image $\inputstyle$, then $\image_{cs}$ will have a style similar to $\inputstyle$:
\begin{equation}
\begin{aligned}
\loss_{contra}^{G}=-\sum_{i=1}^M {\log \frac{\exp({\outputlatent}_i \cdot {\stylelatent}_i/ \tau)}{\exp ({\outputlatent}_i \cdot {\stylelatent}_i / \tau)+\sum_{j=1}^N{ \exp({\outputlatent}_i \cdot {\latentcode_{i_j}^{-}} / \tau)}}},
\end{aligned}
\end{equation}
where $\outputlatent$ and $\stylelatent$ denote the contrastive representation of $\image_{cs}$ and $\inputstyle$, respectively.
The negative examples are sampled from the same dictionary used for training of the MSP module.
Notably, we take the specific generated and reference images as positive examples and utilize contrastive loss as guidance to transfer styles, which is an one-on-one process.
Differently, the contrastive loss in IEST~\cite{chen2021artistic} is calculated only within generated results and it takes a set of images as positive examples, which may reduce the style consistency with the given reference (see Figure~\ref{fig:IECAST_theirs}).

\subsection{Domain Enhancement}
We introduce DE with adversarial loss to enable the network to learn the style distribution
Recent style transfer models employ GAN~\cite{goodfellow2014generative} to align the distribution of generated images with specific artistic images~\cite{chen2021dualast,Lin:2021:DAM}.
The adversarial loss can enhance the holistic style of the stylization results, while it strongly relies on the distribution of datasets.
Even with the specific artistic style loss, the generation process is often not robust enough to be artifact-free.

Differently from these previous methods, we divide the images in the training set into realistic domain and artistic domain, and we use two discriminators $D_R$ and $D_A$ to enhance them respectively (see Figure~\ref{fig:framework}).
During the training process, we first randomly select an image from the realistic domain as the content image $\inputcontent$ and another image from the artistic domain as the style image $\inputstyle$.
$\inputcontent$ and $\inputstyle$ are used as the real samples of $D_R$ and $D_A$, respectively.
The generated image $\image_{cs} = \generator(\inputcontent, \inputstyle)$ is used as the fake sample of $D_A$.
We exchange the content and style images to generate an image $\image_{sc} = \generator(\inputstyle, \inputcontent)$ as the fake sample of $D_R$. The adversarial loss is:
\begin{equation}
\begin{aligned}
\loss_{adv} =& \mathbb{E}[\log D_R(I_c)]+\mathbb{E}[\log (1-D_R(I_{cs}))]\\
& + \mathbb{E}[\log D_A(I_s)]+\mathbb{E}[\log (1-D_A(I_{sc}))],
\end{aligned}
\end{equation}

To maintain the content information of the content image in the process of style transfer between the two domains, we also add a cycle consistency loss:
\begin{equation}
\begin{aligned}
    \loss_{cyc} = \mathbb{E} [\Vert I_c -\generator(I_{cs},I_c)\Vert_1] + \mathbb{E} [\Vert I_s -\generator(I_{sc},I_s)\Vert_1].
    \end{aligned}
\end{equation}

\subsection{Network Training}
Our full objective function for training of the generator $G$ and discriminators $D_R$ and $D_A$ is formulated as:
\begin{equation}
\begin{aligned}
\loss(G, D_R, D_A) &= \lambda_{1} \loss_{adv}+ \lambda_{2}\loss_{cyc}+ \lambda_{3} \loss^G_{contra},\\
\end{aligned}
\end{equation}
where $\lambda_{1}$, $\lambda_{2}$, and $\lambda_{3}$ are weights to balance different loss terms. We set $\lambda_{1} = 1$, $\lambda_{2} = 2$, and $\lambda_{3} = 0.2$ in all of our experiments.

\paragraph{Implementation details}
We collect 100,000 artistic images in different styles from WikiArt~\cite{Phillips:2011:wikiart} and randomly sample 20,000 images as our artistic dataset.
We averagely sample 20,000 images from Places365~\cite{Zhou:2018:Places365} as realistic image dataset.
We train and evaluate our framework on those artistic and realistic images.
In the training phase, all images are loaded with $256 \times 256$ resolution.
The number of feature map layers $M$ is set to be 4.
The dimension $K$ of style latent code is set to 512, 1024, 2048, and 2048 for the four different layers, respectively.
We use Adam~\cite{kingma2014adam} as optimizer with $\beta_1=0.5$, $\beta_2=0.999$, and a batch size of $4$.
The initial learning rate is set to $1 \times 10^{-4}$ and linear decayed linear for total $8 \times 10^5$ iterations.
The training process takes about $18$ hours on one NVIDIA GeForce RTX3090.
We choose the same backbone as AdaIN~\cite{Huang:2017:AdaIn} in our experiments for simplicity.
The results of using other backbones are shown in the supplementary materials.



\section{Experiments}

We compare CAST with several state-of-the-art style transfer methods, including NST~\cite{Gatys:2016:IST}, AdaIN~\cite{Huang:2017:AdaIn}, LST~\cite{Li:2019:LLT}, SANet~\cite{Park:2019:AST}, ArtFlow~\cite{An:2021:Artflow}, MCCNet~\cite{Deng:2021:MCC}, AdaAttN~\cite{liu2021adaattn}, and IEST~\cite{chen2021artistic}.
All the baselines are trained using publicly available implementations with default configurations.
The comparison of inference speed is shown in Table~\ref{table:Quantitative}.


\begin{figure*}[htp]
\newcommand{\galleryfigurewidth}{0.087}
\centering
    \begin{minipage}[t]{\textwidth}
    \centering
        \begin{minipage}{\galleryfigurewidth\linewidth}
        \includegraphics[width=\linewidth]{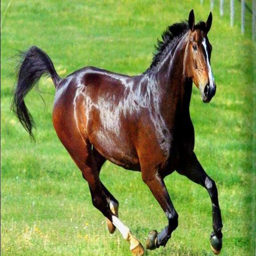}
        \end{minipage}
        \begin{minipage}{\galleryfigurewidth\linewidth}
        \includegraphics[width=\linewidth]{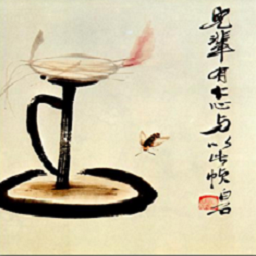}
        \end{minipage}
        \begin{minipage}{\galleryfigurewidth\linewidth}
        \includegraphics[width=\linewidth]{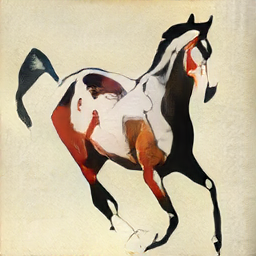}
        \end{minipage}
        \begin{minipage}{\galleryfigurewidth\linewidth}
        \includegraphics[width=\linewidth]{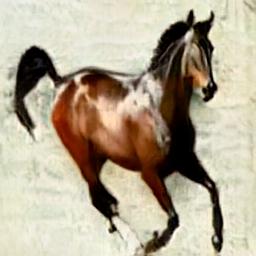}
        \end{minipage}
        \begin{minipage}{\galleryfigurewidth\linewidth}
        \includegraphics[width=\linewidth]{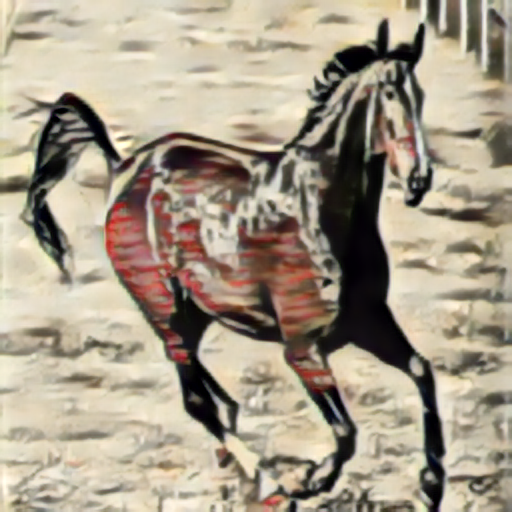}
        \end{minipage}
        \begin{minipage}{\galleryfigurewidth\linewidth}
        \includegraphics[width=\linewidth]{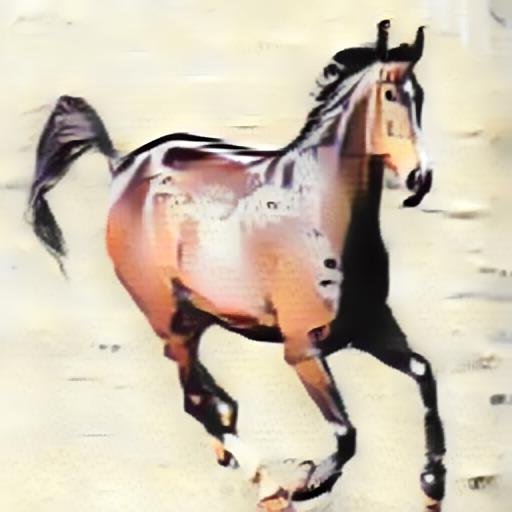}
        \end{minipage}
        \begin{minipage}{\galleryfigurewidth\linewidth}
        \includegraphics[width=\linewidth]{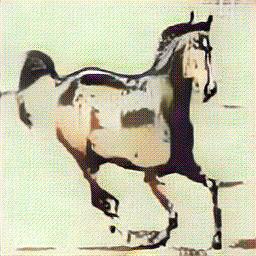}
        \end{minipage}
        \begin{minipage}{\galleryfigurewidth\linewidth}
        \includegraphics[width=\linewidth]{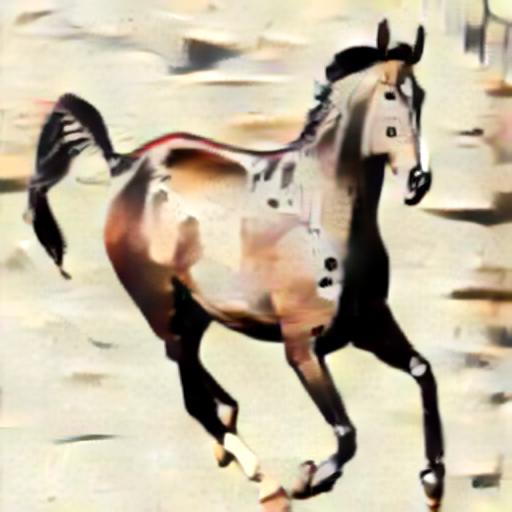}
        \end{minipage}
        \begin{minipage}{\galleryfigurewidth\linewidth}
        \includegraphics[width=\linewidth]{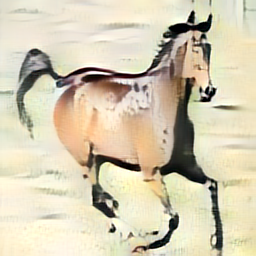}
        \end{minipage}
        \begin{minipage}{\galleryfigurewidth\linewidth}
        \includegraphics[width=\linewidth]{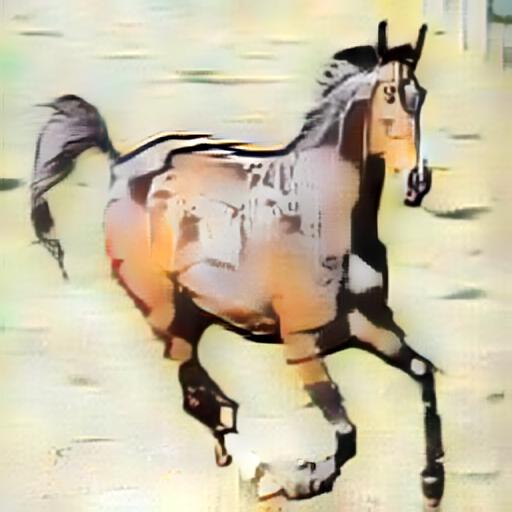}
        \end{minipage}
        \begin{minipage}{\galleryfigurewidth\linewidth}
        \includegraphics[width=\linewidth]{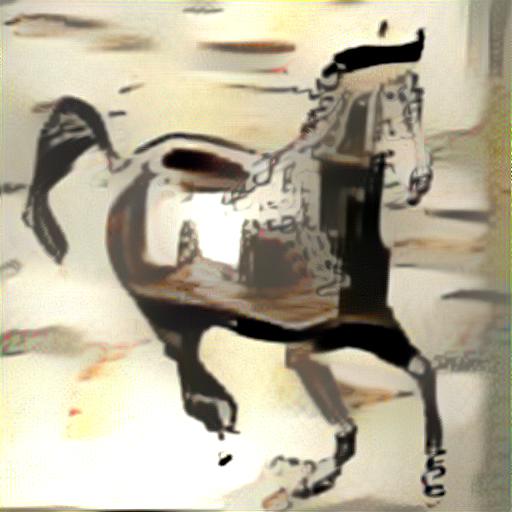}
        \end{minipage}
    \end{minipage}
    \begin{minipage}[t]{\textwidth}
    \centering
        \begin{minipage}{\galleryfigurewidth\linewidth}
        \includegraphics[width=\linewidth]{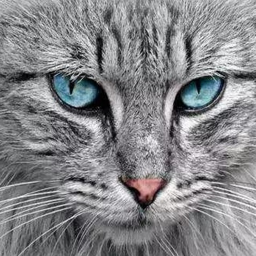}
        \end{minipage}
        \begin{minipage}{\galleryfigurewidth\linewidth}
        \includegraphics[width=\linewidth]{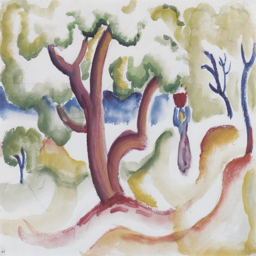}
        \end{minipage}
        \begin{minipage}{\galleryfigurewidth\linewidth}
        \includegraphics[width=\linewidth]{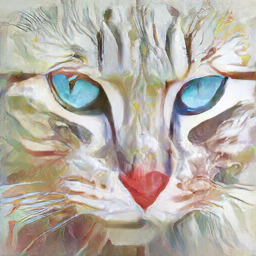}
        \end{minipage}
        \begin{minipage}{\galleryfigurewidth\linewidth}
        \includegraphics[width=\linewidth]{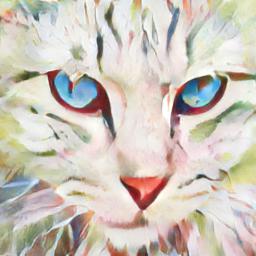}
        \end{minipage}
        \begin{minipage}{\galleryfigurewidth\linewidth}
        \includegraphics[width=\linewidth]{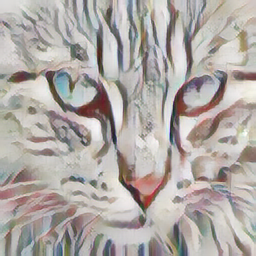}
        \end{minipage}
        \begin{minipage}{\galleryfigurewidth\linewidth}
        \includegraphics[width=\linewidth]{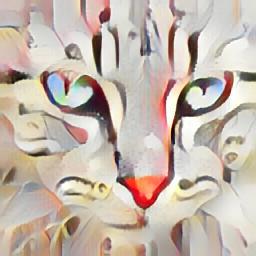}
        \end{minipage}
        \begin{minipage}{\galleryfigurewidth\linewidth}
        \includegraphics[width=\linewidth]{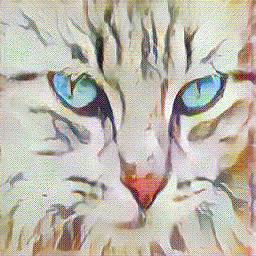}
        \end{minipage}
        \begin{minipage}{\galleryfigurewidth\linewidth}
        \includegraphics[width=\linewidth]{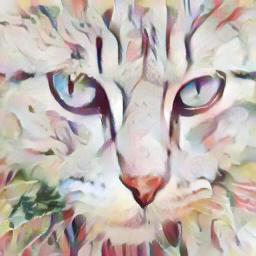}
        \end{minipage}
        \begin{minipage}{\galleryfigurewidth\linewidth}
        \includegraphics[width=\linewidth]{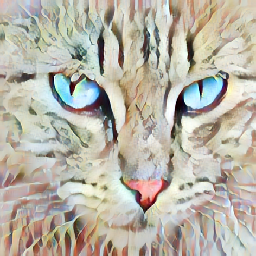}
        \end{minipage}
        \begin{minipage}{\galleryfigurewidth\linewidth}
        \includegraphics[width=\linewidth]{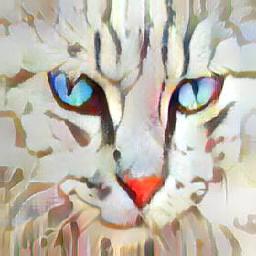}
        \end{minipage}
        \begin{minipage}{\galleryfigurewidth\linewidth}
        \includegraphics[width=\linewidth]{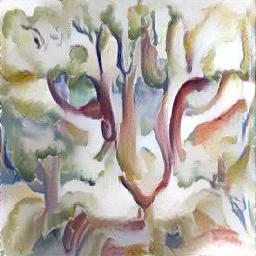}
        \end{minipage}
    \end{minipage}
    \begin{minipage}[t]{\textwidth}
    \centering
        \begin{minipage}{\galleryfigurewidth\linewidth}
        \includegraphics[width=\linewidth]{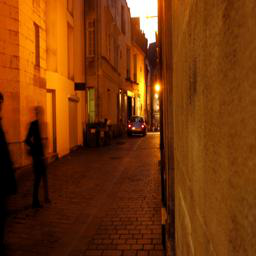}
        \end{minipage}
        \begin{minipage}{\galleryfigurewidth\linewidth}
        \includegraphics[width=\linewidth]{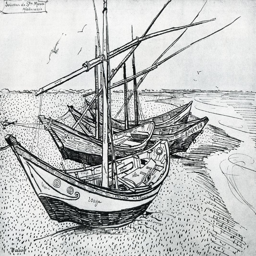}
        \end{minipage}
        \begin{minipage}{\galleryfigurewidth\linewidth}
        \includegraphics[width=\linewidth]{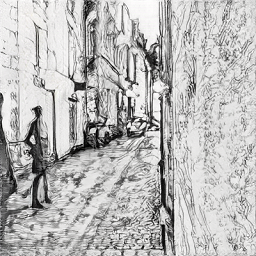}
        \end{minipage}
        \begin{minipage}{\galleryfigurewidth\linewidth}
        \includegraphics[width=\linewidth]{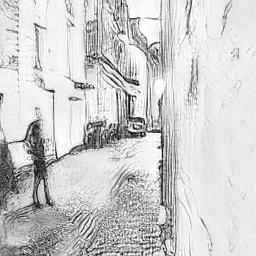}
        \end{minipage}
        \begin{minipage}{\galleryfigurewidth\linewidth}
        \includegraphics[width=\linewidth]{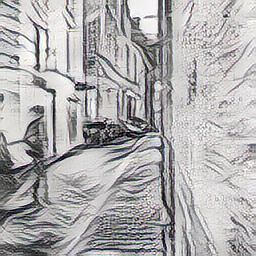}
        \end{minipage}
        \begin{minipage}{\galleryfigurewidth\linewidth}
        \includegraphics[width=\linewidth]{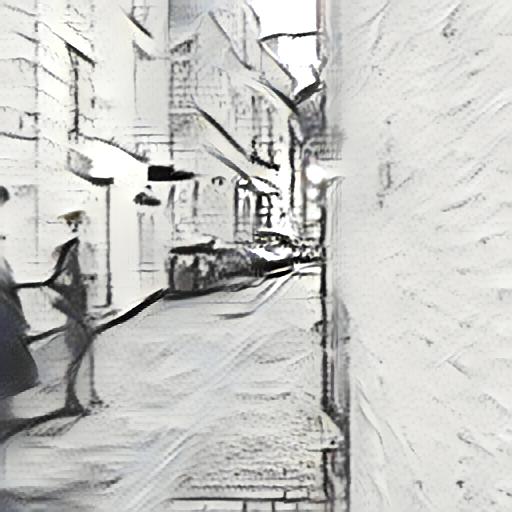}
        \end{minipage}
        \begin{minipage}{\galleryfigurewidth\linewidth}
        \includegraphics[width=\linewidth]{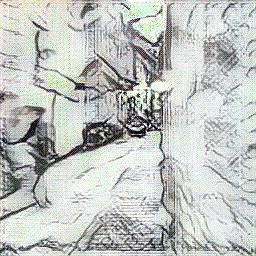}
        \end{minipage}
        \begin{minipage}{\galleryfigurewidth\linewidth}
        \includegraphics[width=\linewidth]{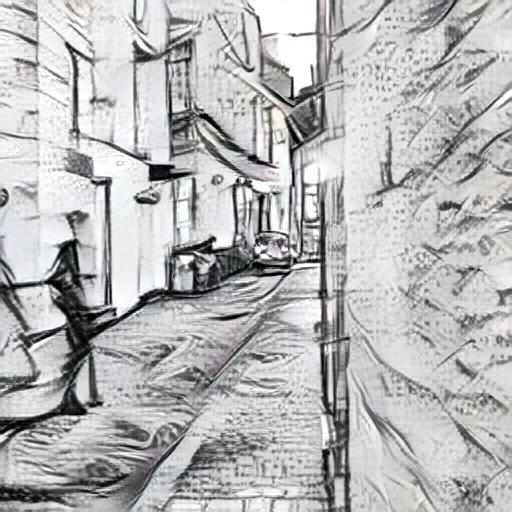}
        \end{minipage}
        \begin{minipage}{\galleryfigurewidth\linewidth}
        \includegraphics[width=\linewidth]{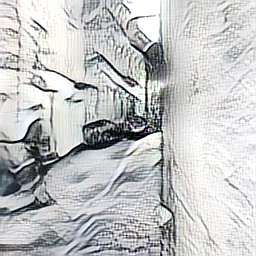}
        \end{minipage}
        \begin{minipage}{\galleryfigurewidth\linewidth}
        \includegraphics[width=\linewidth]{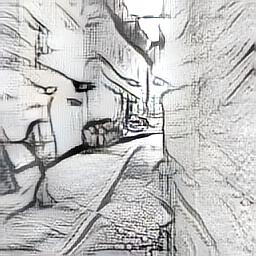}
        \end{minipage}
        \begin{minipage}{\galleryfigurewidth\linewidth}
        \includegraphics[width=\linewidth]{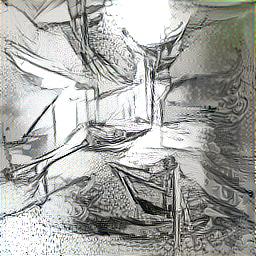}
        \end{minipage}
    \end{minipage}
    \begin{minipage}[t]{\textwidth}
    \centering
        \begin{minipage}{\galleryfigurewidth\linewidth}
        \includegraphics[width=\linewidth]{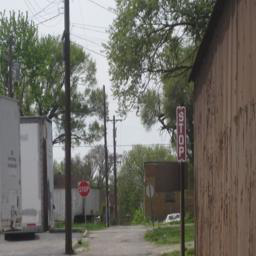}
        \end{minipage}
        \begin{minipage}{\galleryfigurewidth\linewidth}
        \includegraphics[width=\linewidth]{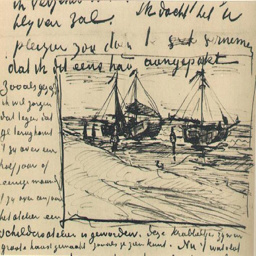}
        \end{minipage}
        \begin{minipage}{\galleryfigurewidth\linewidth}
        \includegraphics[width=\linewidth]{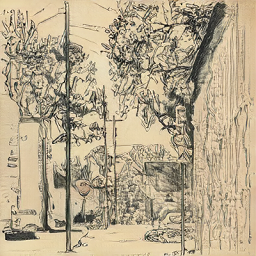}
        \end{minipage}
        \begin{minipage}{\galleryfigurewidth\linewidth}
        \includegraphics[width=\linewidth]{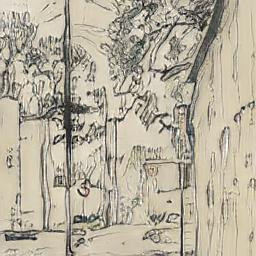}
        \end{minipage}
        \begin{minipage}{\galleryfigurewidth\linewidth}
        \includegraphics[width=\linewidth]{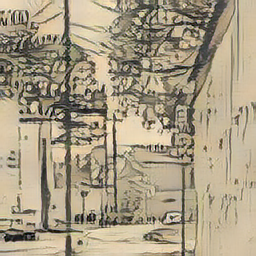}
        \end{minipage}
        \begin{minipage}{\galleryfigurewidth\linewidth}
        \includegraphics[width=\linewidth]{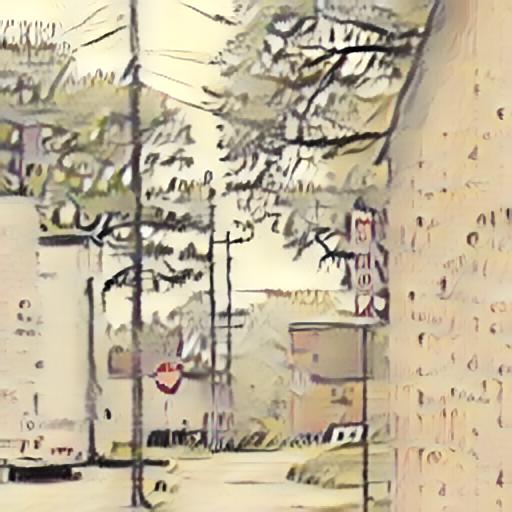}
        \end{minipage}
        \begin{minipage}{\galleryfigurewidth\linewidth}
        \includegraphics[width=\linewidth]{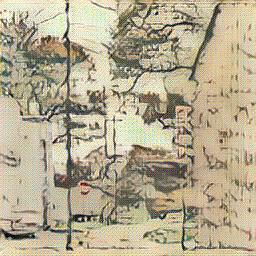}
        \end{minipage}
        \begin{minipage}{\galleryfigurewidth\linewidth}
        \includegraphics[width=\linewidth]{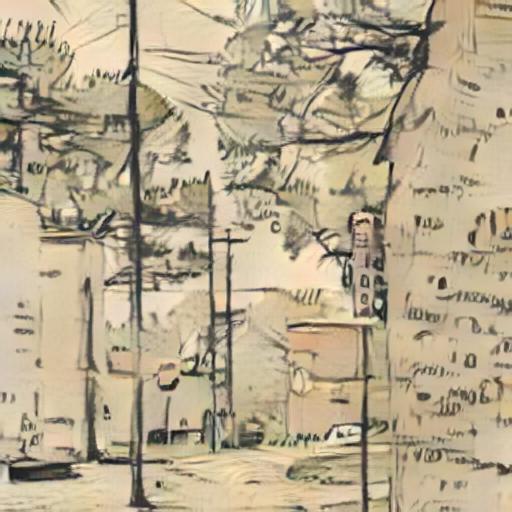}
        \end{minipage}
        \begin{minipage}{\galleryfigurewidth\linewidth}
        \includegraphics[width=\linewidth]{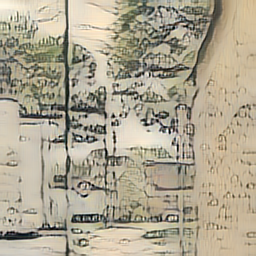}
        \end{minipage}
        \begin{minipage}{\galleryfigurewidth\linewidth}
        \includegraphics[width=\linewidth]{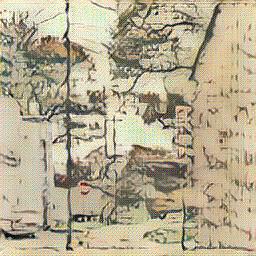}
        \end{minipage}
        \begin{minipage}{\galleryfigurewidth\linewidth}
        \includegraphics[width=\linewidth]{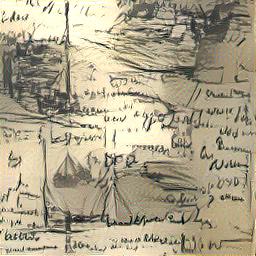}
        \end{minipage}
    \end{minipage}
    \begin{minipage}[t]{\textwidth}
    \centering
        \begin{minipage}{\galleryfigurewidth\linewidth}
        \includegraphics[width=\linewidth]{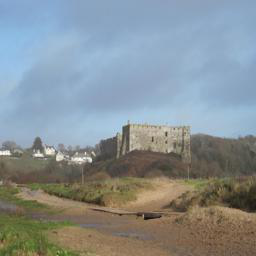}
        \end{minipage}
        \begin{minipage}{\galleryfigurewidth\linewidth}
        \includegraphics[width=\linewidth]{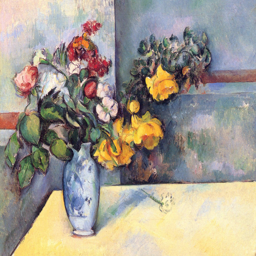}
        \end{minipage}
        \begin{minipage}{\galleryfigurewidth\linewidth}
        \includegraphics[width=\linewidth]{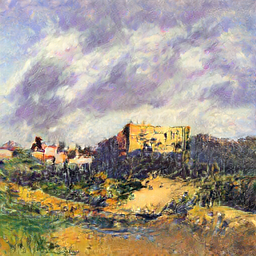}
        \end{minipage}
        \begin{minipage}{\galleryfigurewidth\linewidth}
        \includegraphics[width=\linewidth]{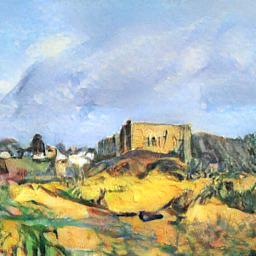}
        \end{minipage}
        \begin{minipage}{\galleryfigurewidth\linewidth}
        \includegraphics[width=\linewidth]{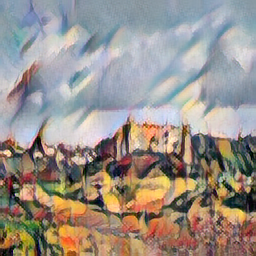}
        \end{minipage}
        \begin{minipage}{\galleryfigurewidth\linewidth}
        \includegraphics[width=\linewidth]{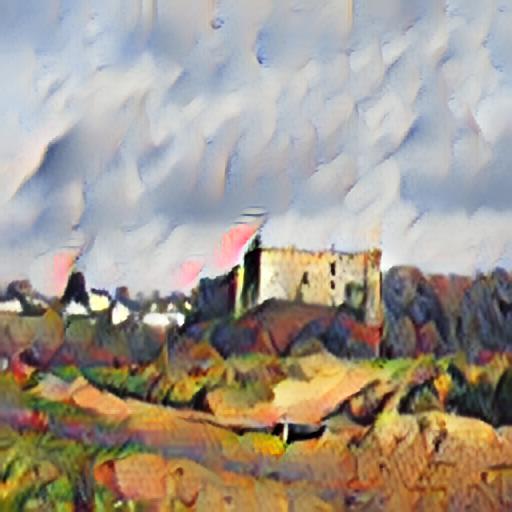}
        \end{minipage}
        \begin{minipage}{\galleryfigurewidth\linewidth}
        \includegraphics[width=\linewidth]{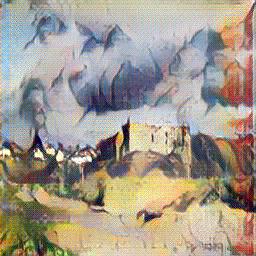}
        \end{minipage}
        \begin{minipage}{\galleryfigurewidth\linewidth}
        \includegraphics[width=\linewidth]{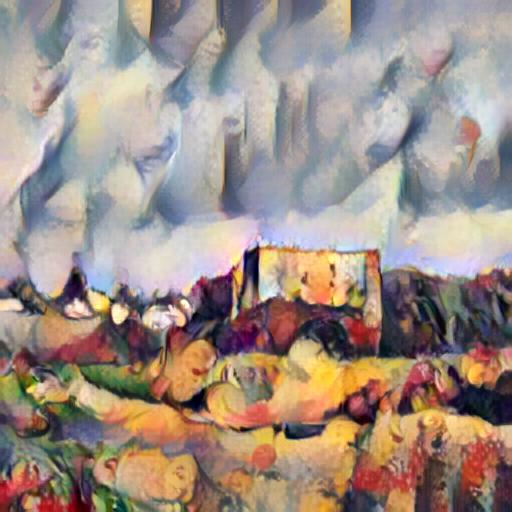}
        \end{minipage}
        \begin{minipage}{\galleryfigurewidth\linewidth}
        \includegraphics[width=\linewidth]{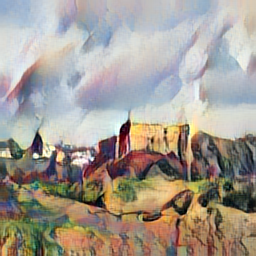}
        \end{minipage}
        \begin{minipage}{\galleryfigurewidth\linewidth}
        \includegraphics[width=\linewidth]{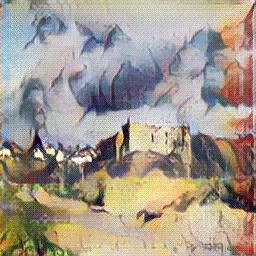}
        \end{minipage}
        \begin{minipage}{\galleryfigurewidth\linewidth}
        \includegraphics[width=\linewidth]{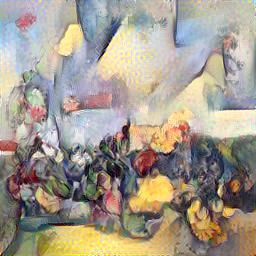}
        \end{minipage}
    \end{minipage}
    \begin{minipage}[t]{\textwidth}
    \centering
        \begin{minipage}{\galleryfigurewidth\linewidth}
        \includegraphics[width=\linewidth]{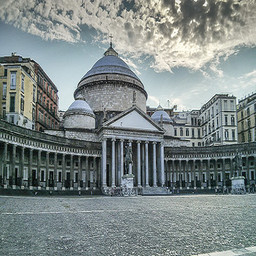}
        \end{minipage}
        \begin{minipage}{\galleryfigurewidth\linewidth}
        \includegraphics[width=\linewidth]{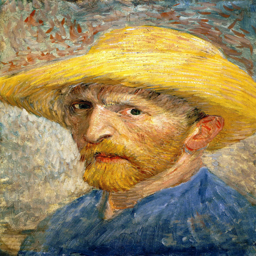}
        \end{minipage}
        \begin{minipage}{\galleryfigurewidth\linewidth}
        \includegraphics[width=\linewidth]{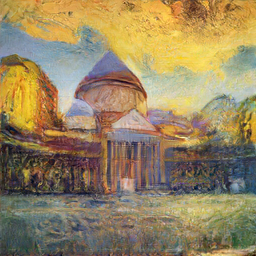}
        \end{minipage}
        \begin{minipage}{\galleryfigurewidth\linewidth}
        \includegraphics[width=\linewidth]{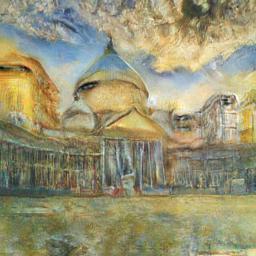}
        \end{minipage}
        \begin{minipage}{\galleryfigurewidth\linewidth}
        \includegraphics[width=\linewidth]{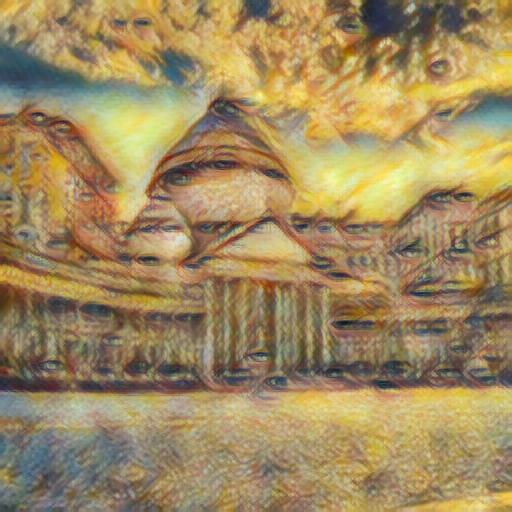}
        \end{minipage}
        \begin{minipage}{\galleryfigurewidth\linewidth}
        \includegraphics[width=\linewidth]{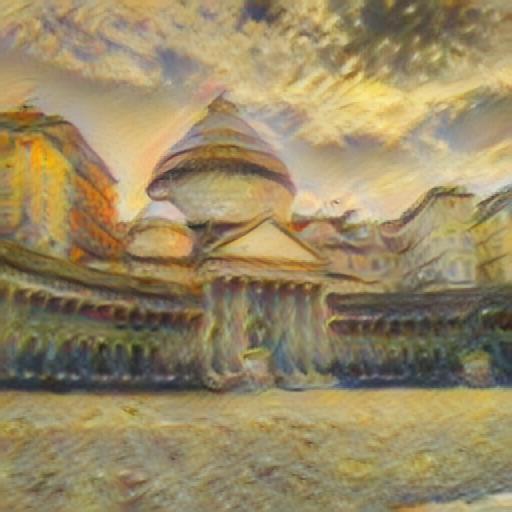}
        \end{minipage}
        \begin{minipage}{\galleryfigurewidth\linewidth}
        \includegraphics[width=\linewidth]{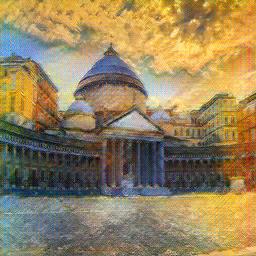}
        \end{minipage}
        \begin{minipage}{\galleryfigurewidth\linewidth}
        \includegraphics[width=\linewidth]{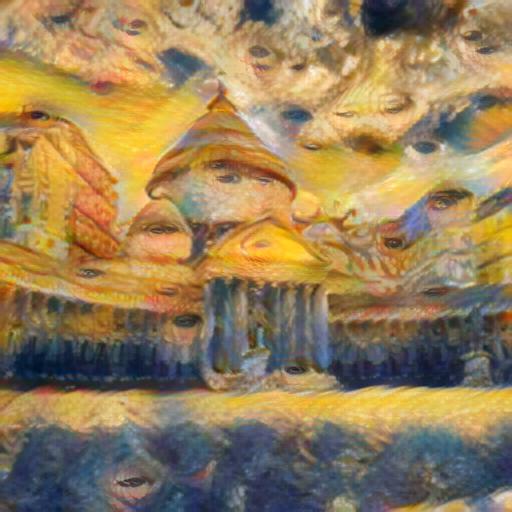}
        \end{minipage}
        \begin{minipage}{\galleryfigurewidth\linewidth}
        \includegraphics[width=\linewidth]{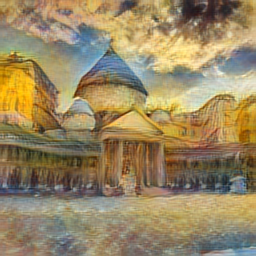}
        \end{minipage}
        \begin{minipage}{\galleryfigurewidth\linewidth}
        \includegraphics[width=\linewidth]{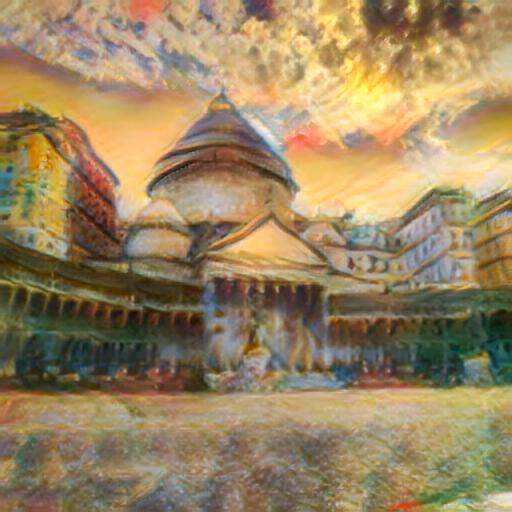}
        \end{minipage}
        \begin{minipage}{\galleryfigurewidth\linewidth}
        \includegraphics[width=\linewidth]{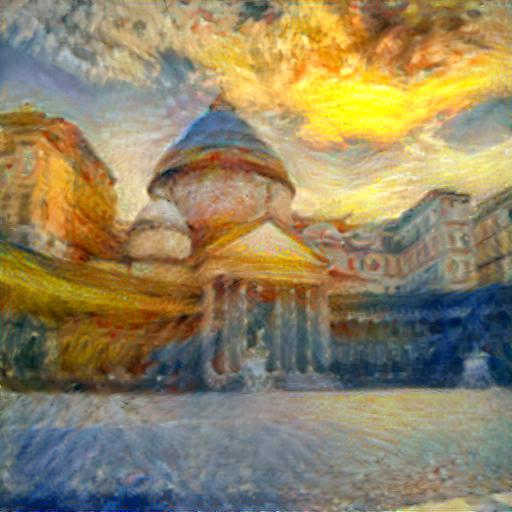}
        \end{minipage}
    \end{minipage}
    \begin{minipage}[t]{\textwidth}
    \centering
        \begin{minipage}{\galleryfigurewidth\linewidth}
        \includegraphics[width=\linewidth]{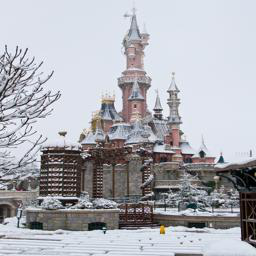}
        \end{minipage}
        \begin{minipage}{\galleryfigurewidth\linewidth}
        \includegraphics[width=\linewidth]{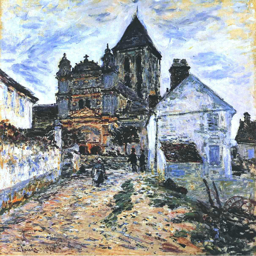}
        \end{minipage}
        \begin{minipage}{\galleryfigurewidth\linewidth}
        \includegraphics[width=\linewidth]{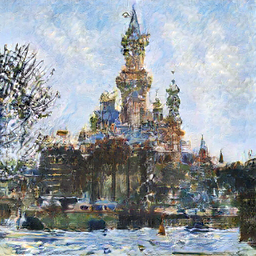}
        \end{minipage}
        \begin{minipage}{\galleryfigurewidth\linewidth}
        \includegraphics[width=\linewidth]{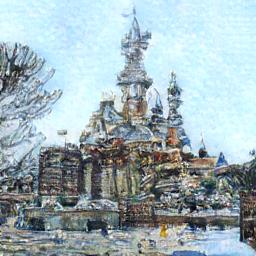}
        \end{minipage}
        \begin{minipage}{\galleryfigurewidth\linewidth}
        \includegraphics[width=\linewidth]{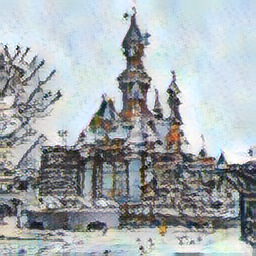}
        \end{minipage}
        \begin{minipage}{\galleryfigurewidth\linewidth}
        \includegraphics[width=\linewidth]{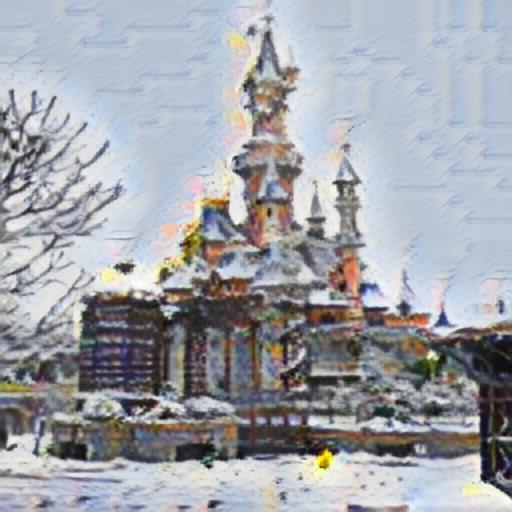}
        \end{minipage}
        \begin{minipage}{\galleryfigurewidth\linewidth}
        \includegraphics[width=\linewidth]{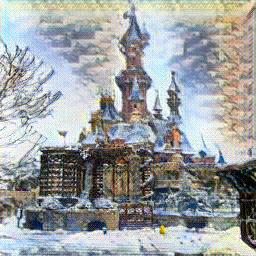}
        \end{minipage}
        \begin{minipage}{\galleryfigurewidth\linewidth}
        \includegraphics[width=\linewidth]{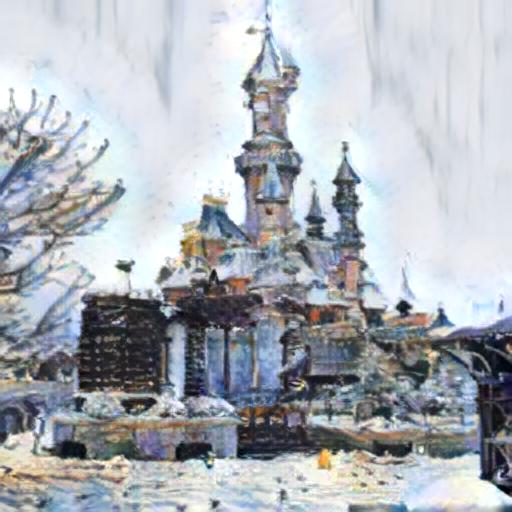}
        \end{minipage}
        \begin{minipage}{\galleryfigurewidth\linewidth}
        \includegraphics[width=\linewidth]{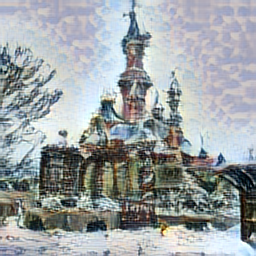}
        \end{minipage}
        \begin{minipage}{\galleryfigurewidth\linewidth}
        \includegraphics[width=\linewidth]{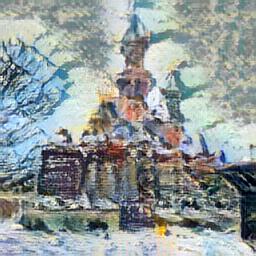}
        \end{minipage}
        \begin{minipage}{\galleryfigurewidth\linewidth}
        \includegraphics[width=\linewidth]{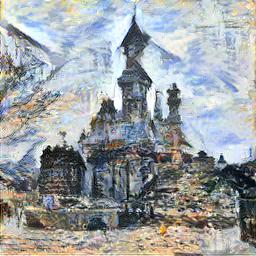}
        \end{minipage}
    \end{minipage}
    \begin{minipage}[t]{\textwidth}
    \centering
        \begin{minipage}{\galleryfigurewidth\linewidth}
        \includegraphics[width=\linewidth]{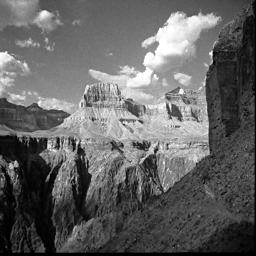}
        \end{minipage}
        \begin{minipage}{\galleryfigurewidth\linewidth}
        \includegraphics[width=\linewidth]{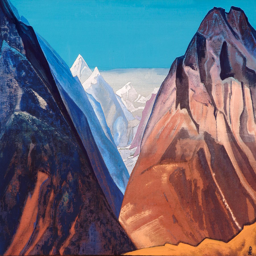}
        \end{minipage}
        \begin{minipage}{\galleryfigurewidth\linewidth}
        \includegraphics[width=\linewidth]{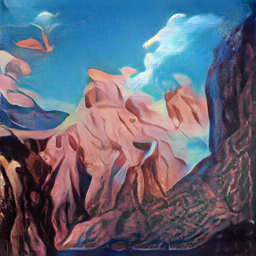}
        \end{minipage}
        \begin{minipage}{\galleryfigurewidth\linewidth}
        \includegraphics[width=\linewidth]{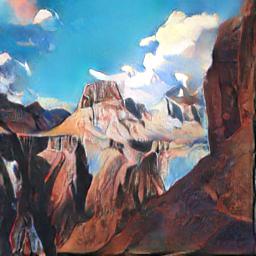}
        \end{minipage}
        \begin{minipage}{\galleryfigurewidth\linewidth}
        \includegraphics[width=\linewidth]{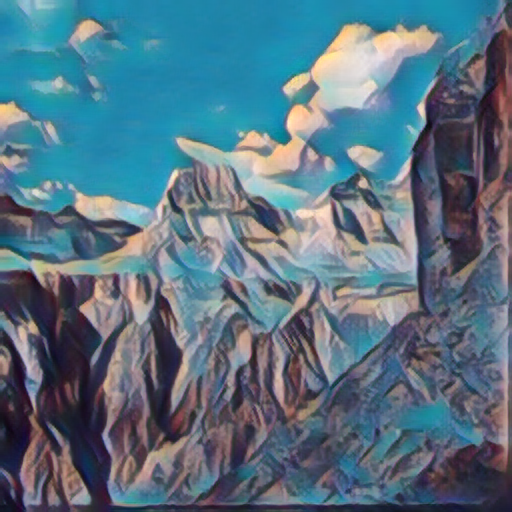}
        \end{minipage}
        \begin{minipage}{\galleryfigurewidth\linewidth}
        \includegraphics[width=\linewidth]{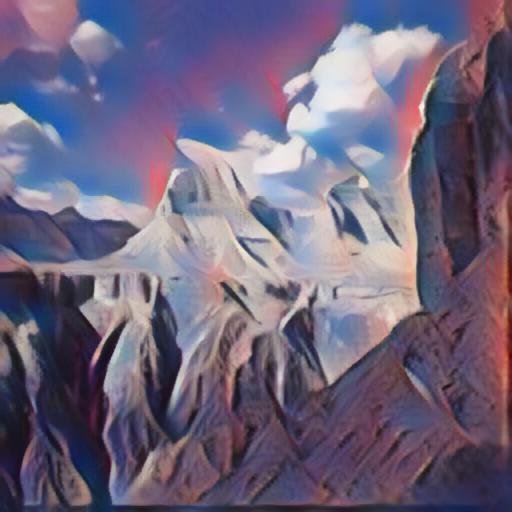}
        \end{minipage}
        \begin{minipage}{\galleryfigurewidth\linewidth}
        \includegraphics[width=\linewidth]{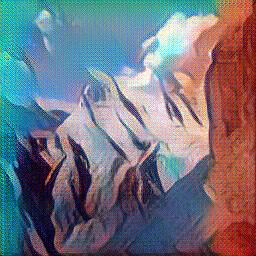}
        \end{minipage}
        \begin{minipage}{\galleryfigurewidth\linewidth}
        \includegraphics[width=\linewidth]{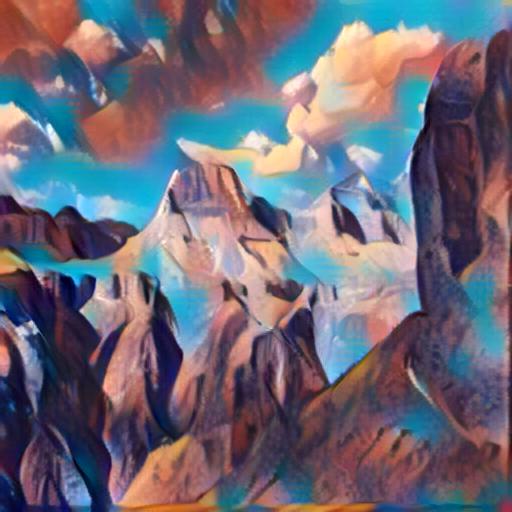}
        \end{minipage}
        \begin{minipage}{\galleryfigurewidth\linewidth}
        \includegraphics[width=\linewidth]{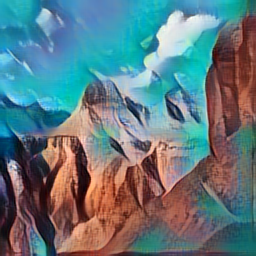}
        \end{minipage}
        \begin{minipage}{\galleryfigurewidth\linewidth}
        \includegraphics[width=\linewidth]{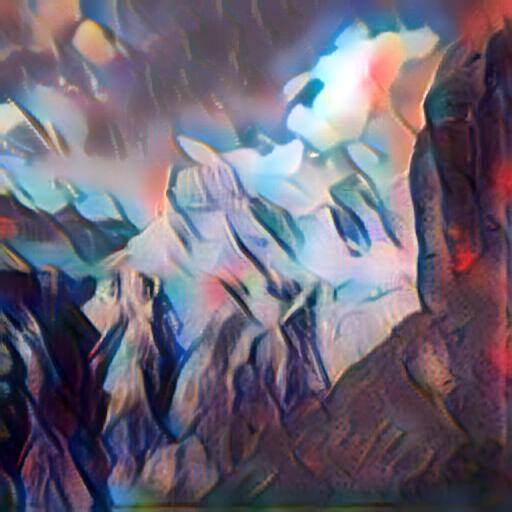}
        \end{minipage}
        \begin{minipage}{\galleryfigurewidth\linewidth}
        \includegraphics[width=\linewidth]{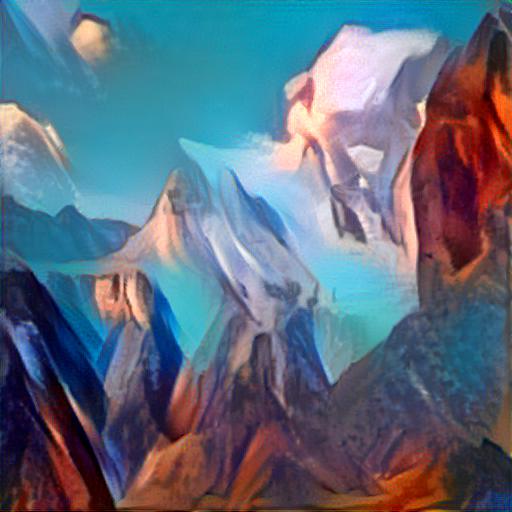}
        \end{minipage}
    \end{minipage}

\begin{subfigure}[t]{\galleryfigurewidth\linewidth}
    \subcaption*{Content}
    \label{fig:SOTA_a}
\end{subfigure}
\begin{subfigure}[t]{\galleryfigurewidth\linewidth}
    \subcaption*{Style}
    \label{fig:SOTA_b}
\end{subfigure}
\begin{subfigure}[t]{\galleryfigurewidth\linewidth}
    \subcaption*{Ours}
    \label{fig:SOTA_c}
\end{subfigure}
\begin{subfigure}[t]{\galleryfigurewidth\linewidth}
    \subcaption*{IEST}
    \label{fig:SOTA_d}
\end{subfigure}
\begin{subfigure}[t]{\galleryfigurewidth\linewidth}
    \subcaption*{AdaAttN}
    \label{fig:SOTA_e}
\end{subfigure}
\begin{subfigure}[t]{\galleryfigurewidth\linewidth}
 \subcaption*{MCCNet}
    \label{fig:SOTA_f}
\end{subfigure}
\begin{subfigure}[t]{\galleryfigurewidth\linewidth}
    \subcaption*{ArtFlow}
    \label{fig:SOTA_g}
\end{subfigure}
\begin{subfigure}[t]{\galleryfigurewidth\linewidth}
    \subcaption*{SANet}
    \label{fig:SOTA_h}
\end{subfigure}
\begin{subfigure}[t]{\galleryfigurewidth\linewidth}
    \subcaption*{LST}
    \label{fig:SOTA_i}
\end{subfigure}
\begin{subfigure}[t]{\galleryfigurewidth\linewidth}
    \subcaption*{AdaIN}
    \label{fig:SOTA_j}
\end{subfigure}
\begin{subfigure}[t]{\galleryfigurewidth\linewidth}
    \subcaption*{NST}
    \label{fig:SOTA_k}
\end{subfigure}
 \vspace{-8pt}
\caption{Qualitative comparisons with several state-of-the-art style transfer methods, including IEST~\cite{chen2021artistic}, AdaAttN~\cite{liu2021adaattn}, MCCNet~\cite{Deng:2021:MCC}, ArtFlow~\cite{An:2021:Artflow}, SANet~\cite{Park:2019:AST}, LST~\cite{Li:2019:LLT}, AdaIN~\cite{Huang:2017:AdaIn}, and NST~\cite{Gatys:2016:IST}.
Artists of style images (from top to bottom): Baishi Qi, August Macke, Vincent van Gogh, Vincent van Gogh, Paul Cezanne, Vincent van Gogh, Claude Monet, and Nicholas Roerich.
}
 \vspace{-8pt}
\label{fig:SOTA}
\end{figure*}


\subsection{Qualitative Evaluation}

We first present qualitative results of our method against the selected state-of-the-art methods in Figure~\ref{fig:SOTA}.
The comparison shows the superiority of CAST in terms of visual quality.
NST is likely to encounter the issue of unpleasant local minimum (e.g., the 1\textsuperscript{st}, 5\textsuperscript{th} and 8\textsuperscript{th} rows).
AdaIN often fails to generate sharp details and introduces undesired patterns that do not exist in style images (e.g., the 1\textsuperscript{st}, 3\textsuperscript{rd}, 6\textsuperscript{th} and 8\textsuperscript{th} rows).
LST tends to transfer low-level style patterns like colors but the local details of strokes are often ignored (e.g., the 2\textsuperscript{nd}-5\textsuperscript{th} rows).
SANet often generates repetitive patterns in the stylized images (e.g., the 2\textsuperscript{nd}, 5\textsuperscript{th},  6\textsuperscript{th} and  8\textsuperscript{th} rows).
ArtFlow sometimes generates unexpected colors or patterns in relatively smooth regions in some cases (e.g., the 1\textsuperscript{st}-5\textsuperscript{th} and 8\textsuperscript{th} rows).
MCCNet can effectively preserve the input content but may fail to capture the stroke details and often generates haloing artifacts around object contours (e.g., the 2\textsuperscript{nd}, 4\textsuperscript{th}, 6\textsuperscript{th}-8\textsuperscript{th} rows).
AdaAttN cannot well capture some stroke patterns (e.g., the  1\textsuperscript{st},  3\textsuperscript{rd}, 4\textsuperscript{th} and 6\textsuperscript{th} rows) and fails to transfer important colors of the style references to the results (the 2\textsuperscript{nd} and  8\textsuperscript{th} rows).
Although the generated visual effects of IEST are of high quality, the usage of second-order statistics as style representation causes color distortion (e.g., the 1\textsuperscript{st} row in Figure~\ref{fig:IECAST_theirs} and the 4\textsuperscript{th} row in Figure~\ref{fig:SOTA}) and cannot capture the detailed stylized patterns (e.g., the regions of sky in the 5\textsuperscript{th} and 7\textsuperscript{th} rows in Figure~\ref{fig:SOTA}).
In particular, these state-of-the-art methods cannot capture the \emph{leaving blank} characteristic of Chinese painting style in the 1\textsuperscript{st} row of Figure~\ref{fig:SOTA} and fail to generate results with a clean background.

In comparison, CAST achieves the best stylization performance that balances characteristics of style patterns and content structures.
Instead of using second-order statistics as a global style descriptor, we use an MSP module for style encoding with the help of a DE module for effective learning of style distribution.
Thus, CAST can flexibly represent vivid local stroke characteristics and the overall appearance while still preserving the content structure.
For instance, as shown in Figures~\ref{fig:teaser}h and \ref{fig:IECAST_ours} (the $2\textsuperscript{nd}$ row), CAST successfully captures the large portion of empty regions in the style images, and it generates a stylization results which have salient objects in the center and blank space around.
The whole vague appearance of the style image (drawn by Claude Monet) in Figures~\ref{fig:teaser}e and \ref{fig:IECAST_ours} (the $1\textsuperscript{st}$ row) is also effectively transferred to the content images.

\subsection{Quantitative Evaluation}
We use the content loss~\cite{Li:2017:UST}, LPIPS~\cite{chen2021artistic}, and deception rate~\cite{Sanakoyeu:2018:SAC} and conduct two user studies to evaluate our method quantitatively.
The two user studies are online surveys that cover art/computer science students/professors and civil servants.

For content loss and LPIPS, we use a pre-trained VGG-19 and compute the average perceptual distances between the content image and the stylized image. The statistics are shown in Table~\ref{tab:user_study}.
For deception rate, we train a VGG-19 network to classify 10 styles on WikiArt.
Then, the deception rate is calculated as the percentage of stylized images that are predicted by the pre-trained network as the correct target styles.
We report the deception rate for the proposed CAST and the baseline models in the 2\textsuperscript{nd} column of Table~\ref{tab:user_study}.
As observed, CAST achieves the highest accuracy and surpasses other methods by a large margin.
As a reference, the mean accuracy of the network on real images of the artists from WikiArt is $78\%$.

\paragraph{User Study \uppercase\expandafter{\romannumeral1}}
We compare CAST with eight state-of-the-art style transfer methods to evaluate which method generates results that are most favored by humans.
For each participant, 50 content-style pairs are randomly selected and the stylized results of CAST and one of the other methods are displayed in a random order.
Then, we ask the participant to choose the image that learns the most characteristics from the style image.
Participants were told that the consistency of content and style was the primary metrics.
The style is subjective and the effectiveness of training also depends on their understanding ability.
Finally, we collect 3,400 votes from 68 participants.
We report the percentage of votes for each method in the 3\textsuperscript{rd} column of Table~\ref{tab:user_study}.
CAST obtains significantly higher preferences in categories of Sketch, Chinese painting, and Impressionism.


\begin{figure*}
\newcommand\ablationfigurewidth{0.107}
\centering
\begin{subfigure}[b]{\ablationfigurewidth\textwidth}
\includegraphics[width=1\linewidth]{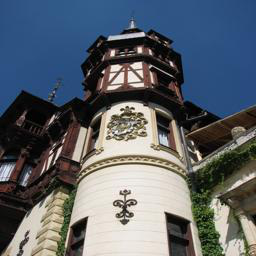}
\includegraphics[width=1\linewidth]{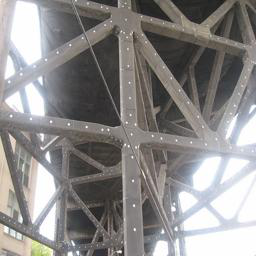}
\caption{Content}
\label{fig:ablation_study_content}
\end{subfigure}
\hskip 0mm
\begin{subfigure}[b]{\ablationfigurewidth\textwidth}
\includegraphics[width=1\linewidth]{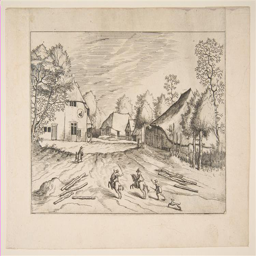}
\includegraphics[width=1\linewidth]{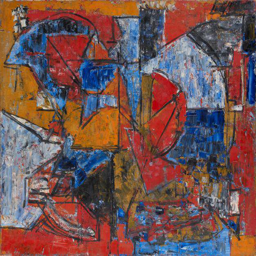}
\caption{Style}
\label{fig:ablation_study_style}
\end{subfigure}
\hskip 0mm
\begin{subfigure}[b]{\ablationfigurewidth\textwidth}
\includegraphics[width=1\linewidth]{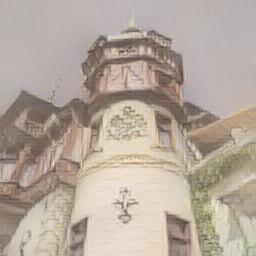}
\includegraphics[width=1\linewidth]{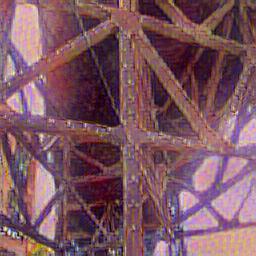}
\caption{AdaIN}
\label{fig:ablation_study_adain}
\end{subfigure}
\hskip 0mm
\begin{subfigure}[b]{\ablationfigurewidth\textwidth}
\includegraphics[width=1\linewidth]{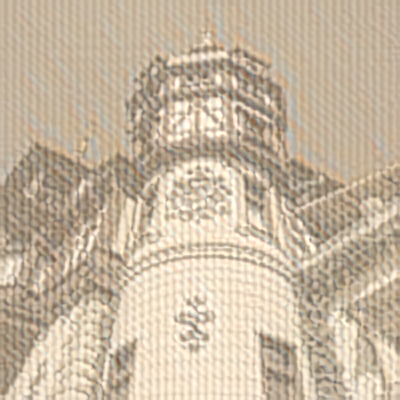}
\includegraphics[width=1\linewidth]{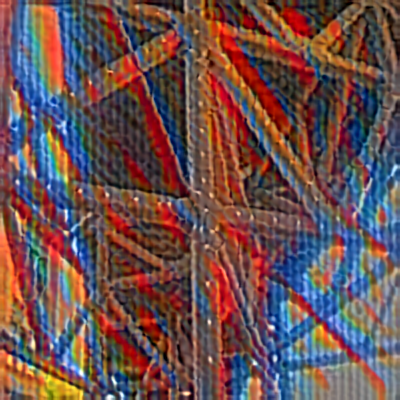}
\caption{w/o DE}
\label{fig:ablation_study_no_domain}
\end{subfigure}
\hskip 0mm
\begin{subfigure}[b]{\ablationfigurewidth\textwidth}
\includegraphics[width=1\linewidth]{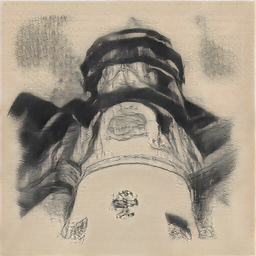}
\includegraphics[width=1\linewidth]{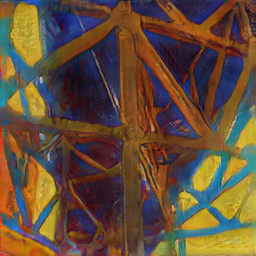}
\caption{w/o $\loss^G_{cont}$}
\label{fig:ablation_study_one_gram}
\end{subfigure}
\hskip 0mm
\begin{subfigure}[b]{\ablationfigurewidth\textwidth}
\includegraphics[width=1\linewidth]{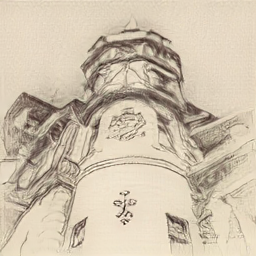}
\includegraphics[width=1\linewidth]{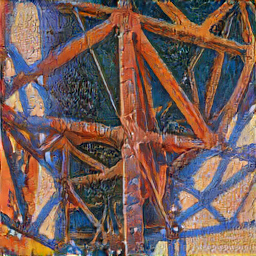}
\caption{mix-DE}
\label{fig:ablation_study_one_domain}
\end{subfigure}
\begin{subfigure}[b]{\ablationfigurewidth\textwidth}
\includegraphics[width=1\linewidth]{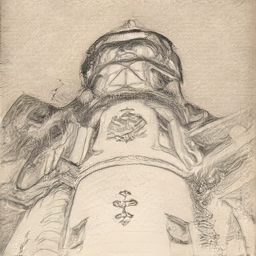}
\includegraphics[width=1\linewidth]{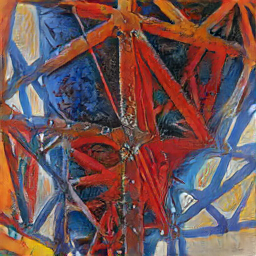}
\caption{one-DE}
\label{fig:ablation_study_art_domain}
\end{subfigure}
\hskip 0mm
\begin{subfigure}[b]{\ablationfigurewidth\textwidth}
\includegraphics[width=1\linewidth]{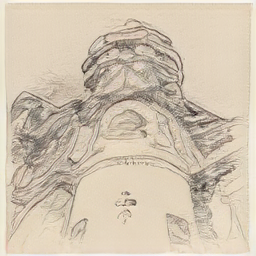}
\includegraphics[width=1\linewidth]{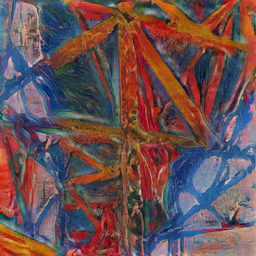}
\caption{$1/2$ cycle}
\label{fig:ablation_study_cyc}
\end{subfigure}
\hskip 0mm
\begin{subfigure}[b]{\ablationfigurewidth\textwidth}
\includegraphics[width=1\linewidth]{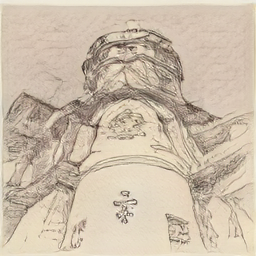}
\includegraphics[width=1\linewidth]{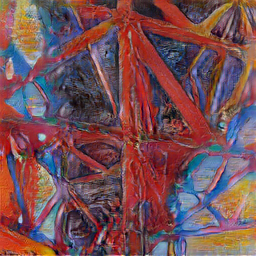}
\caption{full CAST}
\label{fig:ablation_study_full}
\end{subfigure}
\vspace{-8pt}
\caption{Ablation study results.
From left to right: (a) content image; (b) style image; (c) AdaIN with perceptual loss; (d) CAST without DE; (e) AdaIN with style loss based on Gram matrix, DE, and cycle consistency loss; (f) CAST using mixed DE; (g) CAST using one DE without the realistic domain; (h) CAST trained with asymmetric cycle consistent loss by only reconstructing the realistic images; and (i) full CAST model.
Artists of style images: Master of the Small Landscapes and Alexandre Istrati.
}
\label{fig:ablation_study}
\end{figure*}



\begin{table*}
\centering
\caption{Statistics of inference speed and quantitative comparison with state-of-the-art methods.
The results of user study \uppercase\expandafter{\romannumeral1} represent the average percentage of cases in which the result of the corresponding method is preferred compared with ours.
The results of user study \uppercase\expandafter{\romannumeral2} show the accuracy and recall of being selected as fake paintings by the participants.
The best results are in \textbf{bold} while the second best results are marked with \underline{underline}. 
}
\vspace{-8pt}
\begin{tabular}{cc||c|c|c|c|c|c|c|c|c}
\toprule
\multicolumn{2}{c||}{Method} & CAST & IEContraAST & AdaAttN & MCCNet & Artflow & SANet & LST & AdaIN  & NST \\ \hline \hline
\multicolumn{2}{c||}{Inference time (ms/image)$\downarrow$}& \underline{11} & 184 & 130 & 29 & 168 & 14  & \textbf{7} & \underline{11} & 16863  \\
\hline
\multicolumn{2}{c||}{Content loss$\downarrow$} & \underline{0.148} & 0.155 & 0.162 & \textbf{0.117} & 0.172 & 0.150 & 0.155 & 0.176 & 0.188  \\
\hline
\multicolumn{2}{c||}{LPIPS$\downarrow$} & \underline{0.245} &  0.256 & 0.256 & \textbf{0.234} & 0.264 & 0.265 & 0.248 & 0.266 & 0.291\\
\hline
\multicolumn{2}{c||}{Deception Rate$\uparrow$}& \textbf{62.00\%} &  \underline{56.42\%} &   50.70\% &  46.37\% &    43.79\% &  51.87\% &  48.29\% &   51.00\% &  37.70\%   \\
\hline
\multicolumn{2}{c||}{User Study  \uppercase\expandafter{\romannumeral1}} & - &  30.25\% &  41.9\% &  38.3\% &  46.1\% &  44.7\% & 20.0\% &  29.8\% & 25.1\% \\
\hline
\multicolumn{1}{c|}{User Study\uppercase\expandafter{\romannumeral2} }  & Precision$\downarrow$ & \textbf{43.69\%} & 60.26\% & 58.64\% & 70.89\% & \underline{56.81\%} & 65.80\% & 65.49\% & 70.66\% & 60.18\% \\
\cline{2-11}
\multicolumn{1}{c|}{(SAD)} & Recall$\downarrow$ & \textbf{41.19\%} & 58.67\% & \underline{58.16\%} & 72.76\% & 61.55\% & 62.55\% & 64.91\% &75.93& 64.76\%  \\
\bottomrule
\end{tabular}
\label{table:Quantitative}
\label{tab:user_study}
\end{table*}


\paragraph{User Study \uppercase\expandafter{\romannumeral2}}
We design a novel user study to evaluate the stylized images quantitatively, which is called the Stylized Authenticity Detection (SAD).
For each question, we show participants ten artworks of similar styles, including two to four stylized fake painting and ask them to select the synthetic ones.
Within each single question, the stylized paintings are generated by the same method.
Each participant finished 25 questions.
Finally, we collect 2125 groups of results from 85 participants and use the average precision and recall as the measurement for how likely the results will be recognized as synthetics.
Table~\ref{table:Quantitative} shows the statistics.
The paintings generated by CAST have the lowest chance to be decided by people as fake paintings.
We also notice that the precision and recall of CAST is less than $50\%$, which means that users could not tell the real ones from the fakes and tend to select more real paintings as synthetics when doing the testing.

\subsection{Ablation Study}

\paragraph{Contrastive style loss.}
We replace the contrastive style loss with Gram matrix-based perceptual loss, i.e., the model includes perceptual loss, adversarial loss, and cycle consistency loss.
As shown in Figures.~\ref{fig:ablation_study_one_gram} and \ref{fig:ablation_study_full}, the model using Gram matrix instead of our contrastive style loss cannot capture the stroke characteristics of the style image compared with the full CAST model.
The sharp pencil lines of the style image in the 1\textsuperscript{st} row become large black blocks.
The textural oil painting strokes of the style image in the 2\textsuperscript{nd} row become smooth blocks, and the vivid colors becomes murky, while unexpected yellow color appears.
With the contrastive style loss, our full model can faithfully transfer the brushstrokes, textures, and colors from the input style image.

\paragraph{Domain enhancement.}
Our full CAST uses DE for realistic and artistic images separately. We train a simplified CAST model using one discriminator that mixes realistic and artistic images together (mix-DE).
As shown in Figure~\ref{fig:ablation_study_one_domain}, the results generated by mix-DE model are acceptable, but the stroke details in the generated images are weaker than the ones by the full CAST model.
This fact is due to the existence of a significant gap between the artistic and realistic image domains. We further abandon all images from realistic domain for ablation.
As shown in Figure~\ref{fig:ablation_study_art_domain}, the results generated by one-DE model lack details.

\paragraph{Cycle consistency loss.}
To better evaluate the improvement of the contrastive style loss on the style transfer task, we exclude the latent promotion of cycle consistency loss from network training.
The reason is that the reconstruction process of artistic image may imply style information.
We train CAST with an asymmetric cycle consistent loss, which only reconstructs the realistic images.
The decoder of the style transfer network is unaffected by the reconstruction of the artistic image.
As shown in Figure~\ref{fig:ablation_study_cyc}, removing realistic image reconstruction will lead to slightly degraded stylization results.



\section{Conclusion and Future Work}
In this work, we present a novel framework, namely CAST, for the task of arbitrary image style transfer. 
Instead of relying on second-order metrics such as Gram matrix or mean/variance of deep features, we use image features directly by introducing an MSP module for style encoding. 
We develop a contrastive loss function to leverage the available multi-style information in the existing collection of artwork and help train the MSP module and our generative style transfer network.
We further propose a DE scheme to effectively model the distribution of realistic and artistic image domains. 
Extensive experimental results demonstrate that our proposed CAST method achieves superior arbitrary style transfer results compared with state-of-the-art approaches. 
In the future, we plan to improve the contrastive style learning process by considering artist and category information.


\begin{acks}
This work was supported by National Key R\&D Program of China (no. 2020AAA0106200), National Science Foundation of China (nos. 61832016, U20B2070, 6210070958), Ministry of Science and Technology, Taiwan (no. 110-2221-E-006-135-MY3), and Open Projects Program of National Laboratory of Pattern Recognition.
\end{acks}

\bibliographystyle{ACM-Reference-Format}
\bibliography{cast_arxiv.bbl}


\begin{thebibliography}{51}


\ifx \showCODEN    \undefined \def \showCODEN     #1{\unskip}     \fi
\ifx \showDOI      \undefined \def \showDOI       #1{#1}\fi
\ifx \showISBNx    \undefined \def \showISBNx     #1{\unskip}     \fi
\ifx \showISBNxiii \undefined \def \showISBNxiii  #1{\unskip}     \fi
\ifx \showISSN     \undefined \def \showISSN      #1{\unskip}     \fi
\ifx \showLCCN     \undefined \def \showLCCN      #1{\unskip}     \fi
\ifx \shownote     \undefined \def \shownote      #1{#1}          \fi
\ifx \showarticletitle \undefined \def \showarticletitle #1{#1}   \fi
\ifx \showURL      \undefined \def \showURL       {\relax}        \fi
\providecommand\bibfield[2]{#2}
\providecommand\bibinfo[2]{#2}
\providecommand\natexlab[1]{#1}
\providecommand\showeprint[2][]{arXiv:#2}

\bibitem[An et~al\mbox{.}(2021)]%
        {An:2021:Artflow}
\bibfield{author}{\bibinfo{person}{Jie An}, \bibinfo{person}{Siyu Huang},
  \bibinfo{person}{Yibing Song}, \bibinfo{person}{Dejing Dou},
  \bibinfo{person}{Wei Liu}, {and} \bibinfo{person}{Jiebo Luo}.}
  \bibinfo{year}{2021}\natexlab{}.
\newblock \showarticletitle{{ArtFlow}: Unbiased Image Style Transfer via
  Reversible Neural Flows}. In \bibinfo{booktitle}{\emph{IEEE/CVF Conferences
  on Computer Vision and Pattern Recognition (CVPR)}}.
  \bibinfo{pages}{862--871}.
\newblock


\bibitem[Baek et~al\mbox{.}(2021)]%
        {baek2021tunit}
\bibfield{author}{\bibinfo{person}{Kyungjune Baek}, \bibinfo{person}{Yunjey
  Choi}, \bibinfo{person}{Youngjung Uh}, \bibinfo{person}{Jaejun Yoo}, {and}
  \bibinfo{person}{Hyunjung Shim}.} \bibinfo{year}{2021}\natexlab{}.
\newblock \showarticletitle{Rethinking the truly unsupervised image-to-image
  translation}. In \bibinfo{booktitle}{\emph{IEEE/CVF International Conference
  on Computer Vision (ICCV)}}. \bibinfo{pages}{14154--14163}.
\newblock


\bibitem[Chen et~al\mbox{.}(2017)]%
        {chen2017stylebank}
\bibfield{author}{\bibinfo{person}{Dongdong Chen}, \bibinfo{person}{Lu Yuan},
  \bibinfo{person}{Jing Liao}, \bibinfo{person}{Nenghai Yu}, {and}
  \bibinfo{person}{Gang Hua}.} \bibinfo{year}{2017}\natexlab{}.
\newblock \showarticletitle{{StyleBank}: An explicit representation for neural
  image style transfer}. In \bibinfo{booktitle}{\emph{IEEE Conference on
  Computer Vision and Pattern Recognition (CVPR)}}.
  \bibinfo{pages}{1897--1906}.
\newblock


\bibitem[Chen et~al\mbox{.}(2021a)]%
        {chen2021artistic}
\bibfield{author}{\bibinfo{person}{Haibo Chen}, \bibinfo{person}{Lei Zhao},
  \bibinfo{person}{Zhizhong Wang}, \bibinfo{person}{Zhang~Hui Ming},
  \bibinfo{person}{Zhiwen Zuo}, \bibinfo{person}{Ailin Li},
  \bibinfo{person}{Wei Xing}, {and} \bibinfo{person}{Dongming Lu}.}
  \bibinfo{year}{2021}\natexlab{a}.
\newblock \showarticletitle{Artistic Style Transfer with Internal-external
  Learning and Contrastive Learning}. In \bibinfo{booktitle}{\emph{Advances in
  Neural Information Processing Systems (NeurIPS)}}.
\newblock


\bibitem[Chen et~al\mbox{.}(2021b)]%
        {chen2021dualast}
\bibfield{author}{\bibinfo{person}{Haibo Chen}, \bibinfo{person}{Lei Zhao},
  \bibinfo{person}{Zhizhong Wang}, \bibinfo{person}{Huiming Zhang},
  \bibinfo{person}{Zhiwen Zuo}, \bibinfo{person}{Ailin Li},
  \bibinfo{person}{Wei Xing}, {and} \bibinfo{person}{Dongming Lu}.}
  \bibinfo{year}{2021}\natexlab{b}.
\newblock \showarticletitle{{DualAST}: Dual Style-Learning Networks for
  Artistic Style Transfer}. In \bibinfo{booktitle}{\emph{IEEE/CVF Conference on
  Computer Vision and Pattern Recognition (CVPR)}}. \bibinfo{pages}{872--881}.
\newblock


\bibitem[Deng et~al\mbox{.}(2021)]%
        {Deng:2021:MCC}
\bibfield{author}{\bibinfo{person}{Yingying Deng}, \bibinfo{person}{Fan Tang},
  \bibinfo{person}{Weiming Dong}, \bibinfo{person}{Haibin Huang},
  \bibinfo{person}{Chongyang Ma}, {and} \bibinfo{person}{Changsheng Xu}.}
  \bibinfo{year}{2021}\natexlab{}.
\newblock \showarticletitle{Arbitrary Video Style Transfer via Multi-Channel
  Correlation}. In \bibinfo{booktitle}{\emph{AAAI Conference on Artificial
  Intelligence (AAAI)}}. \bibinfo{pages}{1210--1217}.
\newblock


\bibitem[Deng et~al\mbox{.}(2022)]%
        {deng2021stytr2}
\bibfield{author}{\bibinfo{person}{Yingying Deng}, \bibinfo{person}{Fan Tang},
  \bibinfo{person}{Weiming Dong}, \bibinfo{person}{Chongyang Ma},
  \bibinfo{person}{Xingjia Pan}, \bibinfo{person}{Lei Wang}, {and}
  \bibinfo{person}{Changsheng Xu}.} \bibinfo{year}{2022}\natexlab{}.
\newblock \showarticletitle{StyTr$^2$: Image Style Transfer with Transformers}.
  In \bibinfo{booktitle}{\emph{IEEE/CVF Conference on Computer Vision and
  Pattern Recognition (CVPR)}}.
\newblock


\bibitem[Deng et~al\mbox{.}(2020)]%
        {deng2020arbitrary}
\bibfield{author}{\bibinfo{person}{Yingying Deng}, \bibinfo{person}{Fan Tang},
  \bibinfo{person}{Weiming Dong}, \bibinfo{person}{Wen Sun},
  \bibinfo{person}{Feiyue Huang}, {and} \bibinfo{person}{Changsheng Xu}.}
  \bibinfo{year}{2020}\natexlab{}.
\newblock \showarticletitle{Arbitrary style transfer via multi-adaptation
  network}. In \bibinfo{booktitle}{\emph{ACM International Conference on
  Multimedia}}. \bibinfo{pages}{2719--2727}.
\newblock


\bibitem[Dumoulin et~al\mbox{.}(2017)]%
        {dumoulin2016learned}
\bibfield{author}{\bibinfo{person}{Vincent Dumoulin}, \bibinfo{person}{Jonathon
  Shlens}, {and} \bibinfo{person}{Manjunath Kudlur}.}
  \bibinfo{year}{2017}\natexlab{}.
\newblock \showarticletitle{A learned representation for artistic style}. In
  \bibinfo{booktitle}{\emph{International Conference on Learning
  Representations}}.
\newblock


\bibitem[Fi\v{s}er et~al\mbox{.}(2016)]%
        {Fivser:2016:Stylit}
\bibfield{author}{\bibinfo{person}{Jakub Fi\v{s}er},
  \bibinfo{person}{Ond\v{r}ej Jamri\v{s}ka}, \bibinfo{person}{Michal
  Luk\'{a}\v{c}}, \bibinfo{person}{Eli Shechtman}, \bibinfo{person}{Paul
  Asente}, \bibinfo{person}{Jingwan Lu}, {and} \bibinfo{person}{Daniel
  S\'{y}kora}.} \bibinfo{year}{2016}\natexlab{}.
\newblock \showarticletitle{StyLit: Illumination-Guided Example-Based
  Stylization of 3D Renderings}.
\newblock \bibinfo{journal}{\emph{ACM Transactions on Graphics}}
  \bibinfo{volume}{35}, \bibinfo{number}{4}, Article \bibinfo{articleno}{92}
  (\bibinfo{year}{2016}), \bibinfo{numpages}{11}~pages.
\newblock
\showISSN{0730-0301}


\bibitem[Gao et~al\mbox{.}(2020)]%
        {gao2020fast}
\bibfield{author}{\bibinfo{person}{Wei Gao}, \bibinfo{person}{Yijun Li},
  \bibinfo{person}{Yihang Yin}, {and} \bibinfo{person}{Ming-Hsuan Yang}.}
  \bibinfo{year}{2020}\natexlab{}.
\newblock \showarticletitle{Fast video multi-style transfer}. In
  \bibinfo{booktitle}{\emph{IEEE/CVF Winter Conference on Applications of
  Computer Vision}}. \bibinfo{pages}{3222--3230}.
\newblock


\bibitem[Gatys et~al\mbox{.}(2016)]%
        {Gatys:2016:IST}
\bibfield{author}{\bibinfo{person}{Leon~A Gatys}, \bibinfo{person}{Alexander~S
  Ecker}, {and} \bibinfo{person}{Matthias Bethge}.}
  \bibinfo{year}{2016}\natexlab{}.
\newblock \showarticletitle{Image style transfer using convolutional neural
  networks}. In \bibinfo{booktitle}{\emph{IEEE/CVF Conference on Computer
  Vision and Pattern Recognition (CVPR)}}. \bibinfo{pages}{2414--2423}.
\newblock


\bibitem[Gatys et~al\mbox{.}(2017)]%
        {Gatys:2017:CPF}
\bibfield{author}{\bibinfo{person}{Leon~A. Gatys},
  \bibinfo{person}{Alexander~S. Ecker}, \bibinfo{person}{Matthias Bethge},
  \bibinfo{person}{Aaron Hertzmann}, {and} \bibinfo{person}{Eli Shechtman}.}
  \bibinfo{year}{2017}\natexlab{}.
\newblock \showarticletitle{Controlling Perceptual Factors in Neural Style
  Transfer}. In \bibinfo{booktitle}{\emph{IEEE Conference on Computer Vision
  and Pattern Recognition (CVPR)}}. \bibinfo{pages}{3730--3738}.
\newblock


\bibitem[Goodfellow et~al\mbox{.}(2014)]%
        {goodfellow2014generative}
\bibfield{author}{\bibinfo{person}{Ian Goodfellow}, \bibinfo{person}{Jean
  Pouget-Abadie}, \bibinfo{person}{Mehdi Mirza}, \bibinfo{person}{Bing Xu},
  \bibinfo{person}{David Warde-Farley}, \bibinfo{person}{Sherjil Ozair},
  \bibinfo{person}{Aaron Courville}, {and} \bibinfo{person}{Yoshua Bengio}.}
  \bibinfo{year}{2014}\natexlab{}.
\newblock \showarticletitle{Generative Adversarial Nets}. In
  \bibinfo{booktitle}{\emph{Advances in Neural Information Processing Systems
  (NIPS)}}.
\newblock


\bibitem[Han et~al\mbox{.}(2021)]%
        {Han:2021:DCL}
\bibfield{author}{\bibinfo{person}{Junlin Han}, \bibinfo{person}{Mehrdad
  Shoeiby}, \bibinfo{person}{Lars Petersson}, {and}
  \bibinfo{person}{Mohammad~Ali Armin}.} \bibinfo{year}{2021}\natexlab{}.
\newblock \showarticletitle{Dual Contrastive Learning for Unsupervised
  Image-to-Image Translation}. In \bibinfo{booktitle}{\emph{IEEE/CVF Conference
  on Computer Vision and Pattern Recognition Workshops}}.
  \bibinfo{pages}{746--755}.
\newblock


\bibitem[He et~al\mbox{.}(2020)]%
        {He:2019:MOCO}
\bibfield{author}{\bibinfo{person}{Kaiming He}, \bibinfo{person}{Haoqi Fan},
  \bibinfo{person}{Yuxin Wu}, \bibinfo{person}{Saining Xie}, {and}
  \bibinfo{person}{Ross Girshick}.} \bibinfo{year}{2020}\natexlab{}.
\newblock \showarticletitle{Momentum contrast for unsupervised visual
  representation learning}. In \bibinfo{booktitle}{\emph{IEEE/CVF Conference on
  Computer Vision and Pattern Recognition (CVPR)}}.
  \bibinfo{pages}{9729--9738}.
\newblock


\bibitem[Huang and Belongie(2017)]%
        {Huang:2017:AdaIn}
\bibfield{author}{\bibinfo{person}{Xun Huang} {and} \bibinfo{person}{Serge
  Belongie}.} \bibinfo{year}{2017}\natexlab{}.
\newblock \showarticletitle{Arbitrary style transfer in real-time with adaptive
  instance normalization}. In \bibinfo{booktitle}{\emph{IEEE International
  Conference on Computer Vision (ICCV)}}. \bibinfo{pages}{1501--1510}.
\newblock


\bibitem[Jeong and Shin(2021)]%
        {Jeong:2021:Contrad}
\bibfield{author}{\bibinfo{person}{Jongheon Jeong} {and}
  \bibinfo{person}{Jinwoo Shin}.} \bibinfo{year}{2021}\natexlab{}.
\newblock \showarticletitle{Training {GAN}s with Stronger Augmentations via
  Contrastive Discriminator}. In \bibinfo{booktitle}{\emph{International
  Conference on Learning Representations}}.
\newblock


\bibitem[Jing et~al\mbox{.}(2020a)]%
        {jing2020dynamic}
\bibfield{author}{\bibinfo{person}{Yongcheng Jing}, \bibinfo{person}{Xiao Liu},
  \bibinfo{person}{Yukang Ding}, \bibinfo{person}{Xinchao Wang},
  \bibinfo{person}{Errui Ding}, \bibinfo{person}{Mingli Song}, {and}
  \bibinfo{person}{Shilei Wen}.} \bibinfo{year}{2020}\natexlab{a}.
\newblock \showarticletitle{Dynamic instance normalization for arbitrary style
  transfer}. In \bibinfo{booktitle}{\emph{Proceedings of the AAAI Conference on
  Artificial Intelligence}}. \bibinfo{pages}{4369--4376}.
\newblock


\bibitem[Jing et~al\mbox{.}(2020b)]%
        {Jing:2020:NSTReview}
\bibfield{author}{\bibinfo{person}{Yongcheng Jing}, \bibinfo{person}{Yezhou
  Yang}, \bibinfo{person}{Zunlei Feng}, \bibinfo{person}{Jingwen Ye},
  \bibinfo{person}{Yizhou Yu}, {and} \bibinfo{person}{Mingli Song}.}
  \bibinfo{year}{2020}\natexlab{b}.
\newblock \showarticletitle{Neural Style Transfer: A Review}.
\newblock \bibinfo{journal}{\emph{IEEE Transactions on Visualization and
  Computer Graphics}} \bibinfo{volume}{26}, \bibinfo{number}{11}
  (\bibinfo{year}{2020}), \bibinfo{pages}{3365--3385}.
\newblock


\bibitem[Johnson et~al\mbox{.}(2016)]%
        {johnson2016perceptual}
\bibfield{author}{\bibinfo{person}{Justin Johnson}, \bibinfo{person}{Alexandre
  Alahi}, {and} \bibinfo{person}{Li Fei-Fei}.} \bibinfo{year}{2016}\natexlab{}.
\newblock \showarticletitle{Perceptual losses for real-time style transfer and
  super-resolution}. In \bibinfo{booktitle}{\emph{European Conference on
  Computer Vision (ECCV)}}. Springer, \bibinfo{pages}{694--711}.
\newblock


\bibitem[Kang and Park(2020)]%
        {Kang:2020:ContraGAN}
\bibfield{author}{\bibinfo{person}{Minguk Kang} {and} \bibinfo{person}{Jaesik
  Park}.} \bibinfo{year}{2020}\natexlab{}.
\newblock \showarticletitle{ContraGAN: Contrastive Learning for Conditional
  Image Generation}. In \bibinfo{booktitle}{\emph{Advances in Neural
  Information Processing Systems (NeurIPS)}}.
\newblock


\bibitem[Kingma and Ba(2014)]%
        {kingma2014adam}
\bibfield{author}{\bibinfo{person}{Diederik~P Kingma} {and}
  \bibinfo{person}{Jimmy Ba}.} \bibinfo{year}{2014}\natexlab{}.
\newblock \showarticletitle{Adam: A method for stochastic optimization}.
\newblock \bibinfo{journal}{\emph{arXiv preprint arXiv:1412.6980}}
  (\bibinfo{year}{2014}).
\newblock


\bibitem[Kolkin et~al\mbox{.}(2019)]%
        {Kolkin:2019:STR}
\bibfield{author}{\bibinfo{person}{Nicholas Kolkin}, \bibinfo{person}{Jason
  Salavon}, {and} \bibinfo{person}{Gregory Shakhnarovich}.}
  \bibinfo{year}{2019}\natexlab{}.
\newblock \showarticletitle{Style Transfer by Relaxed Optimal Transport and
  Self-Similarity}. In \bibinfo{booktitle}{\emph{IEEE/CVF Conference on
  Computer Vision and Pattern Recognition (CVPR)}}.
  \bibinfo{pages}{10043--10052}.
\newblock


\bibitem[Kotovenko et~al\mbox{.}(2019a)]%
        {kotovenko2019content}
\bibfield{author}{\bibinfo{person}{Dmytro Kotovenko}, \bibinfo{person}{Artsiom
  Sanakoyeu}, \bibinfo{person}{Sabine Lang}, {and} \bibinfo{person}{Bjorn
  Ommer}.} \bibinfo{year}{2019}\natexlab{a}.
\newblock \showarticletitle{Content and style disentanglement for artistic
  style transfer}. In \bibinfo{booktitle}{\emph{IEEE/CVF International
  Conference on Computer Vision (ICCV)}}. \bibinfo{pages}{4422--4431}.
\newblock


\bibitem[Kotovenko et~al\mbox{.}(2019b)]%
        {kotovenko2019content_transformation}
\bibfield{author}{\bibinfo{person}{Dmytro Kotovenko}, \bibinfo{person}{Artsiom
  Sanakoyeu}, \bibinfo{person}{Pingchuan Ma}, \bibinfo{person}{Sabine Lang},
  {and} \bibinfo{person}{Bjorn Ommer}.} \bibinfo{year}{2019}\natexlab{b}.
\newblock \showarticletitle{A content transformation block for image style
  transfer}. In \bibinfo{booktitle}{\emph{IEEE/CVF Conference on Computer
  Vision and Pattern Recognition (CVPR)}}. \bibinfo{pages}{10032--10041}.
\newblock


\bibitem[Li et~al\mbox{.}(2019)]%
        {Li:2019:LLT}
\bibfield{author}{\bibinfo{person}{Xueting Li}, \bibinfo{person}{Sifei Liu},
  \bibinfo{person}{Jan Kautz}, {and} \bibinfo{person}{Ming-Hsuan Yang}.}
  \bibinfo{year}{2019}\natexlab{}.
\newblock \showarticletitle{Learning Linear Transformations for Fast Image and
  Video Style Transfer}. In \bibinfo{booktitle}{\emph{IEEE/CVF Conference on
  Computer Vision and Pattern Recognition (CVPR)}}.
  \bibinfo{pages}{3804--3812}.
\newblock


\bibitem[Li et~al\mbox{.}(2017)]%
        {Li:2017:UST}
\bibfield{author}{\bibinfo{person}{Yijun Li}, \bibinfo{person}{Chen Fang},
  \bibinfo{person}{Jimei Yang}, \bibinfo{person}{Zhaowen Wang},
  \bibinfo{person}{Xin Lu}, {and} \bibinfo{person}{Ming-Hsuan Yang}.}
  \bibinfo{year}{2017}\natexlab{}.
\newblock \showarticletitle{Universal style transfer via feature transforms}.
  In \bibinfo{booktitle}{\emph{Advances Neural Information Processing Systems
  (NeurIPS)}}. \bibinfo{pages}{386--396}.
\newblock


\bibitem[Liao et~al\mbox{.}(2017)]%
        {Liao:2017:VAT}
\bibfield{author}{\bibinfo{person}{Jing Liao}, \bibinfo{person}{Yuan Yao},
  \bibinfo{person}{Lu Yuan}, \bibinfo{person}{Gang Hua}, {and}
  \bibinfo{person}{Sing~Bing Kang}.} \bibinfo{year}{2017}\natexlab{}.
\newblock \showarticletitle{Visual Attribute Transfer through Deep Image
  Analogy}.
\newblock \bibinfo{journal}{\emph{ACM Transactions on Graphics}}
  \bibinfo{volume}{36}, \bibinfo{number}{4}, Article \bibinfo{articleno}{120}
  (\bibinfo{year}{2017}), \bibinfo{numpages}{15}~pages.
\newblock


\bibitem[Lin et~al\mbox{.}(2021)]%
        {Lin:2021:DAM}
\bibfield{author}{\bibinfo{person}{Minxuan Lin}, \bibinfo{person}{Fan Tang},
  \bibinfo{person}{Weiming Dong}, \bibinfo{person}{Xiao Li},
  \bibinfo{person}{Changsheng Xu}, {and} \bibinfo{person}{Chongyang Ma}.}
  \bibinfo{year}{2021}\natexlab{}.
\newblock \showarticletitle{Distribution Aligned Multimodal and Multi-Domain
  Image Stylization}.
\newblock \bibinfo{journal}{\emph{ACM Transactions on Multimedia Computing,
  Communications, and Applications}} \bibinfo{volume}{17}, \bibinfo{number}{3},
  Article \bibinfo{articleno}{96} (\bibinfo{year}{2021}),
  \bibinfo{numpages}{17}~pages.
\newblock
\showISSN{1551-6857}


\bibitem[Liu et~al\mbox{.}(2021a)]%
        {Liu:2021:DivCo}
\bibfield{author}{\bibinfo{person}{Rui Liu}, \bibinfo{person}{Yixiao Ge},
  \bibinfo{person}{Ching~Lam Choi}, \bibinfo{person}{Xiaogang Wang}, {and}
  \bibinfo{person}{Hongsheng Li}.} \bibinfo{year}{2021}\natexlab{a}.
\newblock \showarticletitle{{DivCo}: Diverse Conditional Image Synthesis via
  Contrastive Generative Adversarial Network}. In
  \bibinfo{booktitle}{\emph{IEEE/CVF Conference on Computer Vision and Pattern
  Recognition (CVPR)}}. \bibinfo{pages}{16372--16381}.
\newblock


\bibitem[Liu et~al\mbox{.}(2021b)]%
        {liu2021adaattn}
\bibfield{author}{\bibinfo{person}{Songhua Liu}, \bibinfo{person}{Tianwei Lin},
  \bibinfo{person}{Dongliang He}, \bibinfo{person}{Fu Li},
  \bibinfo{person}{Meiling Wang}, \bibinfo{person}{Xin Li},
  \bibinfo{person}{Zhengxing Sun}, \bibinfo{person}{Qian Li}, {and}
  \bibinfo{person}{Errui Ding}.} \bibinfo{year}{2021}\natexlab{b}.
\newblock \showarticletitle{{AdaAttN}: Revisit attention mechanism in arbitrary
  neural style transfer}. In \bibinfo{booktitle}{\emph{IEEE/CVF International
  Conference on Computer Vision (ICCV)}}. \bibinfo{pages}{6649--6658}.
\newblock


\bibitem[Liu et~al\mbox{.}(2019)]%
        {Liu:2019:EUD}
\bibfield{author}{\bibinfo{person}{Xialei Liu}, \bibinfo{person}{Joost van~de
  Weijer}, {and} \bibinfo{person}{Andrew~D. Bagdanov}.}
  \bibinfo{year}{2019}\natexlab{}.
\newblock \showarticletitle{Exploiting Unlabeled Data in CNNs by
  Self-Supervised Learning to Rank}.
\newblock \bibinfo{journal}{\emph{IEEE Transactions on Pattern Analysis and
  Machine Intelligence}} \bibinfo{volume}{41}, \bibinfo{number}{8}
  (\bibinfo{year}{2019}), \bibinfo{pages}{1862--1878}.
\newblock


\bibitem[Park and Lee(2019)]%
        {Park:2019:AST}
\bibfield{author}{\bibinfo{person}{Dae~Young Park} {and}
  \bibinfo{person}{Kwang~Hee Lee}.} \bibinfo{year}{2019}\natexlab{}.
\newblock \showarticletitle{Arbitrary Style Transfer With Style-Attentional
  Networks}. In \bibinfo{booktitle}{\emph{IEEE/CVF Conference on Computer
  Vision and Pattern Recognition (CVPR)}}. \bibinfo{pages}{5880--5888}.
\newblock


\bibitem[Park et~al\mbox{.}(2020)]%
        {park2020CUT}
\bibfield{author}{\bibinfo{person}{Taesung Park}, \bibinfo{person}{Alexei~A
  Efros}, \bibinfo{person}{Richard Zhang}, {and} \bibinfo{person}{Jun-Yan
  Zhu}.} \bibinfo{year}{2020}\natexlab{}.
\newblock \showarticletitle{Contrastive learning for unpaired image-to-image
  translation}. In \bibinfo{booktitle}{\emph{European Conference on Computer
  Vision (ECCV)}}. Springer, \bibinfo{pages}{319--345}.
\newblock


\bibitem[Phillips and Mackintosh(2011)]%
        {Phillips:2011:wikiart}
\bibfield{author}{\bibinfo{person}{Fred Phillips} {and} \bibinfo{person}{Brandy
  Mackintosh}.} \bibinfo{year}{2011}\natexlab{}.
\newblock \showarticletitle{Wiki Art Gallery, Inc.: A case for critical
  thinking}.
\newblock \bibinfo{journal}{\emph{Issues in Accounting Education}}
  \bibinfo{volume}{26}, \bibinfo{number}{3} (\bibinfo{year}{2011}),
  \bibinfo{pages}{593--608}.
\newblock


\bibitem[Puy and P{\'e}rez(2019)]%
        {puy2019flexible}
\bibfield{author}{\bibinfo{person}{Gilles Puy} {and} \bibinfo{person}{Patrick
  P{\'e}rez}.} \bibinfo{year}{2019}\natexlab{}.
\newblock \showarticletitle{A flexible convolutional solver for fast style
  transfers}. In \bibinfo{booktitle}{\emph{IEEE/CVF Conference on Computer
  Vision and Pattern Recognition (CVPR)}}. \bibinfo{pages}{8963--8972}.
\newblock


\bibitem[Sanakoyeu et~al\mbox{.}(2018a)]%
        {sanakoyeu2018style}
\bibfield{author}{\bibinfo{person}{Artsiom Sanakoyeu}, \bibinfo{person}{Dmytro
  Kotovenko}, \bibinfo{person}{Sabine Lang}, {and} \bibinfo{person}{Bjorn
  Ommer}.} \bibinfo{year}{2018}\natexlab{a}.
\newblock \showarticletitle{A style-aware content loss for real-time {HD} style
  transfer}. In \bibinfo{booktitle}{\emph{European Conference on Computer
  Vision (ECCV)}}. \bibinfo{pages}{698--714}.
\newblock


\bibitem[Sanakoyeu et~al\mbox{.}(2018b)]%
        {Sanakoyeu:2018:SAC}
\bibfield{author}{\bibinfo{person}{Artsiom Sanakoyeu}, \bibinfo{person}{Dmytro
  Kotovenko}, \bibinfo{person}{Sabine Lang}, {and} \bibinfo{person}{Bj{\"o}rn
  Ommer}.} \bibinfo{year}{2018}\natexlab{b}.
\newblock \showarticletitle{A Style-Aware Content Loss for Real-Time {HD} Style
  Transfer}. In \bibinfo{booktitle}{\emph{European Conference Computer Vision
  (ECCV)}}. \bibinfo{publisher}{Springer International Publishing},
  \bibinfo{address}{Cham}, \bibinfo{pages}{715--731}.
\newblock


\bibitem[Santa~Cruz et~al\mbox{.}(2019)]%
        {Cruz:2019:VPL}
\bibfield{author}{\bibinfo{person}{Rodrigo Santa~Cruz}, \bibinfo{person}{Basura
  Fernando}, \bibinfo{person}{Anoop Cherian}, {and} \bibinfo{person}{Stephen
  Gould}.} \bibinfo{year}{2019}\natexlab{}.
\newblock \showarticletitle{Visual Permutation Learning}.
\newblock \bibinfo{journal}{\emph{IEEE Transactions on Pattern Analysis and
  Machine Intelligence}} \bibinfo{volume}{41}, \bibinfo{number}{12}
  (\bibinfo{year}{2019}), \bibinfo{pages}{3100--3114}.
\newblock


\bibitem[Simonyan and Zisserman(2014)]%
        {simonyan2014very}
\bibfield{author}{\bibinfo{person}{Karen Simonyan} {and}
  \bibinfo{person}{Andrew Zisserman}.} \bibinfo{year}{2014}\natexlab{}.
\newblock \showarticletitle{Very deep convolutional networks for large-scale
  image recognition}. In \bibinfo{booktitle}{\emph{International Conference on
  Learning Representations}}.
\newblock


\bibitem[Svoboda et~al\mbox{.}(2020)]%
        {svoboda2020two}
\bibfield{author}{\bibinfo{person}{Jan Svoboda}, \bibinfo{person}{Asha
  Anoosheh}, \bibinfo{person}{Christian Osendorfer}, {and}
  \bibinfo{person}{Jonathan Masci}.} \bibinfo{year}{2020}\natexlab{}.
\newblock \showarticletitle{Two-stage peer-regularized feature recombination
  for arbitrary image style transfer}. In \bibinfo{booktitle}{\emph{IEEE/CVF
  Conference on Computer Vision and Pattern Recognition}}.
  \bibinfo{pages}{13816--13825}.
\newblock


\bibitem[Ulyanov et~al\mbox{.}(2016)]%
        {ulyanov2016texture}
\bibfield{author}{\bibinfo{person}{Dmitry Ulyanov}, \bibinfo{person}{Vadim
  Lebedev}, \bibinfo{person}{Andrea Vedaldi}, {and} \bibinfo{person}{Victor~S
  Lempitsky}.} \bibinfo{year}{2016}\natexlab{}.
\newblock \showarticletitle{Texture networks: Feed-forward synthesis of
  textures and stylized images}. In \bibinfo{booktitle}{\emph{International
  Conference on Machine Learning (ICML)}}. \bibinfo{pages}{1349–1357}.
\newblock


\bibitem[Van~den Oord et~al\mbox{.}(2018)]%
        {van2018representation}
\bibfield{author}{\bibinfo{person}{Aaron Van~den Oord}, \bibinfo{person}{Yazhe
  Li}, \bibinfo{person}{Oriol Vinyals}, {et~al\mbox{.}}}
  \bibinfo{year}{2018}\natexlab{}.
\newblock \showarticletitle{Representation learning with contrastive predictive
  coding}.
\newblock \bibinfo{journal}{\emph{arXiv preprint arXiv:1807.03748}}
  \bibinfo{volume}{2}, \bibinfo{number}{3} (\bibinfo{year}{2018}),
  \bibinfo{pages}{4}.
\newblock


\bibitem[Wang et~al\mbox{.}(2004)]%
        {Wang:2004:EEP}
\bibfield{author}{\bibinfo{person}{Bin Wang}, \bibinfo{person}{Wenping Wang},
  \bibinfo{person}{Huaiping Yang}, {and} \bibinfo{person}{Jiaguang Sun}.}
  \bibinfo{year}{2004}\natexlab{}.
\newblock \showarticletitle{Efficient example-based painting and synthesis of
  {2D} directional texture}.
\newblock \bibinfo{journal}{\emph{IEEE Transactions on Visualization and
  Computer Graphics}} \bibinfo{volume}{10}, \bibinfo{number}{3}
  (\bibinfo{year}{2004}), \bibinfo{pages}{266--277}.
\newblock


\bibitem[Wu et~al\mbox{.}(2021b)]%
        {Wu:2021:CLC}
\bibfield{author}{\bibinfo{person}{Haiyan Wu}, \bibinfo{person}{Yanyun Qu},
  \bibinfo{person}{Shaohui Lin}, \bibinfo{person}{Jian Zhou},
  \bibinfo{person}{Ruizhi Qiao}, \bibinfo{person}{Zhizhong Zhang},
  \bibinfo{person}{Yuan Xie}, {and} \bibinfo{person}{Lizhuang Ma}.}
  \bibinfo{year}{2021}\natexlab{b}.
\newblock \showarticletitle{Contrastive Learning for Compact Single Image
  Dehazing}. In \bibinfo{booktitle}{\emph{IEEE/CVF Conference on Computer
  Vision and Pattern Recognition (CVPR)}}. \bibinfo{pages}{10546--10555}.
\newblock


\bibitem[Wu et~al\mbox{.}(2021a)]%
        {Wu:2021:Styleformer}
\bibfield{author}{\bibinfo{person}{Xiaolei Wu}, \bibinfo{person}{Zhihao Hu},
  \bibinfo{person}{Lu Sheng}, {and} \bibinfo{person}{Dong Xu}.}
  \bibinfo{year}{2021}\natexlab{a}.
\newblock \showarticletitle{{StyleFormer}: Real-Time Arbitrary Style Transfer
  via Parametric Style Composition}. In \bibinfo{booktitle}{\emph{IEEE/CVF
  International Conference on Computer Vision (ICCV)}}.
  \bibinfo{pages}{14618--14627}.
\newblock


\bibitem[Xu et~al\mbox{.}(2021)]%
        {xu2021drb}
\bibfield{author}{\bibinfo{person}{Wenju Xu}, \bibinfo{person}{Chengjiang
  Long}, \bibinfo{person}{Ruisheng Wang}, {and} \bibinfo{person}{Guanghui
  Wang}.} \bibinfo{year}{2021}\natexlab{}.
\newblock \showarticletitle{{DRB-GAN}: A Dynamic ResBlock Generative
  Adversarial Network for Artistic Style Transfer}. In
  \bibinfo{booktitle}{\emph{IEEE/CVF International Conference on Computer
  Vision (ICCV)}}. \bibinfo{pages}{6383--6392}.
\newblock


\bibitem[Zhang and Dana(2018)]%
        {zhang2018multi}
\bibfield{author}{\bibinfo{person}{Hang Zhang} {and} \bibinfo{person}{Kristin
  Dana}.} \bibinfo{year}{2018}\natexlab{}.
\newblock \showarticletitle{Multi-style generative network for real-time
  transfer}. In \bibinfo{booktitle}{\emph{European Conference on Computer
  Vision Workshops}}. \bibinfo{pages}{349--365}.
\newblock


\bibitem[Zhou et~al\mbox{.}(2018)]%
        {Zhou:2018:Places365}
\bibfield{author}{\bibinfo{person}{Bolei Zhou}, \bibinfo{person}{Agata
  Lapedriza}, \bibinfo{person}{Aditya Khosla}, \bibinfo{person}{Aude Oliva},
  {and} \bibinfo{person}{Antonio Torralba}.} \bibinfo{year}{2018}\natexlab{}.
\newblock \showarticletitle{Places: A 10 Million Image Database for Scene
  Recognition}.
\newblock \bibinfo{journal}{\emph{IEEE Transactions on Pattern Analysis and
  Machine Intelligence}} \bibinfo{volume}{40}, \bibinfo{number}{6}
  (\bibinfo{year}{2018}), \bibinfo{pages}{1452--1464}.
\newblock


\bibitem[Zhu et~al\mbox{.}(2017)]%
        {Zhu:2017:CycleGAN}
\bibfield{author}{\bibinfo{person}{Jun-Yan Zhu}, \bibinfo{person}{Taesung
  Park}, \bibinfo{person}{Phillip Isola}, {and} \bibinfo{person}{Alexei~A
  Efros}.} \bibinfo{year}{2017}\natexlab{}.
\newblock \showarticletitle{Unpaired image-to-image translation using
  cycle-consistent adversarial networks}. In \bibinfo{booktitle}{\emph{IEEE
  International Conference on Computer Vision (ICCV)}}.
  \bibinfo{pages}{2223--2232}.
\newblock


\end{thebibliography}

\end{document}